\documentclass{article}
\PassOptionsToPackage{numbers, compress}{natbib}
\usepackage[preprint,main]{neurips_2026}

\usepackage[utf8]{inputenc}
\usepackage[T1]{fontenc}
\usepackage{url}
\usepackage{booktabs}
\usepackage{amsfonts}
\usepackage{amsmath}
\usepackage{amssymb}
\usepackage{amsthm}
\usepackage{mathtools}
\usepackage{bm}
\usepackage{nicefrac}
\usepackage{microtype}
\usepackage{xcolor}
\usepackage{graphicx}
\usepackage{enumitem}
\usepackage{algorithm}
\usepackage{algpseudocode}
\usepackage[hidelinks]{hyperref}
\hypersetup{pdftitle={Random Voronoi Networks}}

\title{Random Voronoi Networks}

\author{%
  Ray Verma$^{1}$
  \And Pau Mateo-Bernardó$^{2}$
  \And Daniel Mas Montserrat$^{3,4}$
  \AND
  David Bonet$^{3,4,5}$
  \And Marçal Comajoan Cara$^{3,6}$
  \And Alexander G. Ioannidis$^{3,4}$
  \AND
  \parbox{0.94\textwidth}{\centering\normalfont\small
    $^{1}$New York University\quad
    $^{2}$Universitat Politècnica de Catalunya\\
    $^{3}$Stanford University\quad
    $^{4}$UC Santa Cruz Genomics Institute\\
    $^{5}$Universitat de Barcelona\quad
    $^{6}$UC Berkeley}
}

\newtheorem{theorem}{Theorem}[section]
\newtheorem{proposition}[theorem]{Proposition}
\newtheorem{corollary}[theorem]{Corollary}

\theoremstyle{remark}

\newcommand{\R}{\mathbb{R}}
\newcommand{\E}{\mathbb{E}}
\newcommand{\Prob}{\mathbb{P}}
\newcommand{\1}{\mathbf{1}}

\newcommand{\step}{\mathrm{step}}
\newcommand{\relu}{\mathrm{ReLU}}

\DeclareMathOperator*{\argmin}{arg\,min}
\DeclareMathOperator*{\argmax}{arg\,max}

\begin{document}
\maketitle

\begin{abstract}
Nearest-neighbor methods are local, simple, and interpretable, but their non-parametric form makes them memory-intensive, expensive at inference, and difficult to integrate into differentiable neural architectures. We introduce \emph{Random Voronoi Networks} (RVNs), a family of structured three-layer MLPs that express nearest-neighbor geometry through sampled Voronoi comparisons. The first layer computes \emph{Random Voronoi Features} (RVF) from affine pairwise separators between selected training samples, the second layer aggregates these comparisons into \emph{Voronoi cells} or top-$k$ neighborhood scores, and the readout layer produces classification or regression predictions. We show that, when all samples and pairwise comparisons are included, step RVNs recover exact Euclidean $k$-NN, while ReLU RVNs provide a violation-score realization for 1-NN that can be relaxed into a differentiable soft weighting. Subsampling yields approximate RVNs with explicit control over the number of cells and features. We study several unsupervised sampling methods that trade off random coverage, directional diversity, coordinate sparsity, and empirical feature variance. Empirically, RVFs produce a highly effective data encoding strategy, providing performance boosts to downstream methods ranging from linear to modern tabular deep learning models.  Finally, we showcase that RVF provides an interpretable feature map due to its $k$-NN based construction and ties to pairwise local directions. These results position our Random Voronoi formulation as a bridge between nearest-neighbor rules, random features, and modern tabular models.
\end{abstract}

\section{Introduction}

Nearest-neighbor rules require very few parametric assumptions, adapt naturally to local structure, and come with classical statistical consistency guarantees \citep{cover1967nearest,stone1977consistent,devroye1996probabilistic}. At the same time, their strengths are tied to properties that modern deep learning pipelines often find inconvenient: prediction requires access to the reference set, memory and latency scale with the number of stored examples, and the winner-take-all nearest-neighbor operator is not differentiable. Furthermore, the curse of dimensionality can degrade nearest-neighbor performance in high dimensions, while the unrestricted binary $k$-NN classification class has infinite VC dimension, ruling out distribution-free uniform generalization bounds based solely on its class complexity.

The core object behind $1$-NN is the \emph{Voronoi tessellation} induced by the training set. Prediction is constant inside each cell and changes only across Voronoi boundaries \citep{okabe1992spatial,sitawarin2021voronoi}, which are defined as a combination of hyperplanes. This observation suggests a natural parallelism between \emph{Voronoi tessellation} and \emph{multi-layer perceptrons} (MLPs), since both construct decision regions by composing affine separators with nonlinear aggregation.
This paper develops that viewpoint through \emph{Random Voronoi Networks} (RVNs). RVNs are a family of structured three-layer networks that recover
$1$-NN when every training sample is used as an anchor and each anchor
is compared with every other training sample. The first layer generates \emph{Voronoi features} by applying sampled pairwise Voronoi separators derived from the training set. The second layer generates \emph{Voronoi cells} by aggregating these comparisons onto cell-membership or top-$k$ neighborhood indicators. Finally, a \emph{readout layer} provides the classification or regression prediction. RVN can be instantiated both using step and ReLU activations, inheriting the local inductive bias of prototype methods while remaining compatible with standard neural tooling.

Our contributions are as follows:
(1) We formulate Random Voronoi Networks as structured three-layer MLPs whose layers correspond to Voronoi features, Voronoi cells, and readout predictions.
(2) We show that step RVNs recover exact Euclidean $k$-NN in the full-sample, full-comparison regime, while ReLU RVNs provide a neural violation-score realization of the $1$-NN geometry.
(3) We analyze uniform comparison sampling and define feasible, label-agnostic selectors using random coverage, directional diversity, coordinate sparsity, and empirical feature variance. The structured selectors are evaluated empirically and do not inherit the uniform-sampling bounds.
(4) We empirically separate the exact nearest-neighbor recovery role of RVN from the practical feature-encoding role of RVF, showing improvements over standard random features, comparisons with learned neighborhood models, and interpretable local feature directions.

Together, these results position RVNs as a bridge between nearest-neighbor rules, random feature maps, and neural networks: they preserve the locality and interpretability of prototype methods while offering fixed, scalable, and neural-compatible feature representations.

\section{Related work}

\textbf{Nearest neighbors and prototype methods.}
The classical analysis of nearest-neighbor rules begins with the seminal result of \citet{cover1967nearest} and the consistency theory of \citet{stone1977consistent}; a broader treatment appears in the monograph of \citet{devroye1996probabilistic}. Prototype-based methods such as learning vector quantization (LVQ) compress the reference set into a smaller number of representatives and therefore induce Voronoi decision regions by design \citep{baras1990convergence,sato1996glvq}. More recent prototype literature emphasizes interpretability, robustness, and efficient certification \citep{saralajew2019pbnn,voracek2022robust}. RVNs share the local geometric bias of these methods, but use sampled pairwise Voronoi comparisons as an explicit intermediate representation.

\textbf{Neural Voronoi and $k$-NN implementations.}
Neural implementations of nearest-neighbor and Voronoi decision rules predate this work: Murphy's $1$-NN perceptron uses bisecting-plane comparisons \citep{murphy1990nearest}, Bose and Garga construct an exact Voronoi network \citep{bose1993neural}, Chen et al. implement $k$-NN through an iterative procedure \citep{chen1997neural}, and Gentile and Sznaier simplify Voronoi classifiers using feedback and hyperborder structure \citep{gentile2001improved,gentile2002hyperborders}. 
RVN instead emphasizes a static feedforward construction with explicit anchor and comparison budgets $(M_C,M_F)$, structured comparison subsampling, and reuse of the selected comparisons as a fixed label-agnostic feature map for downstream predictors.

\textbf{Differentiable and learned neighborhood models.}
A complementary line of work learns or smooths nearest-neighbor retrieval so that locality can be inserted inside larger differentiable pipelines. Examples include neural nearest-neighbor blocks \citep{plotz2018n3}, TabR \citep{gorishniy2024tabr}, and ModernNCA \citep{ye2025modernnca}. TabR and ModernNCA learn task-specific representations and neighborhood behavior through specialized training objectives or architectures. RVF instead fixes a Euclidean, label-agnostic comparison map before supervised training and can be attached to linear models, generic MLPs, or specialized tabular predictors without changing their losses. We therefore view RVF as complementary to learned retrieval methods rather than as a replacement for them.

\textbf{Generalization bounds for nearest-neighbor limits.}
A related theoretical line studies why nearest-neighbor rules can generalize despite their large unrestricted capacity. \citet{graepel2021pacbayesian} derive PAC-Bayesian bounds for $K$-NN by viewing it as the limiting case of a kernel classifier with vanishing bandwidth, while \citet{chaudhuri2014rates} give finite-sample, distribution-dependent convergence rates for nearest-neighbor classification. Our perspective is complementary: rather than deriving $k$-NN from a kernel classifier, RVNs realize the same limiting nearest-neighbor geometry through a structured MLP built from Voronoi comparisons, which allows us to characterize how finite comparison budgets approximate nearest-neighbor cells and pairwise rank geometry.

\textbf{Random features and random ReLU networks.}
RVNs also connect to the large literature on random features, where nonlinear dictionaries are sampled and only a simple downstream head is trained \citep{rahimi2007random,hsu2021random}. Unlike random Fourier or random ReLU features, Voronoi features are data-dependent: each coordinate is tied to a specific pairwise separator between reference samples. The resulting representation therefore carries an explicit local and interpretable geometry while retaining the computational benefits of a fixed feature map.

\textbf{Voronoi geometry of $k$-NN.}
The geometric structure of nearest-neighbor classifiers can be analyzed via Voronoi diagrams and their higher-order variants \citep{okabe1992spatial,sitawarin2021voronoi}. We use this geometry constructively: pairwise bisectors become affine features, sampled rank tests become cells, and the complete comparison graph provides the exact $k$-NN endpoint around which the budgeted approximation is defined.

\begin{figure}[t]
    \centering
    \includegraphics[width=\linewidth]{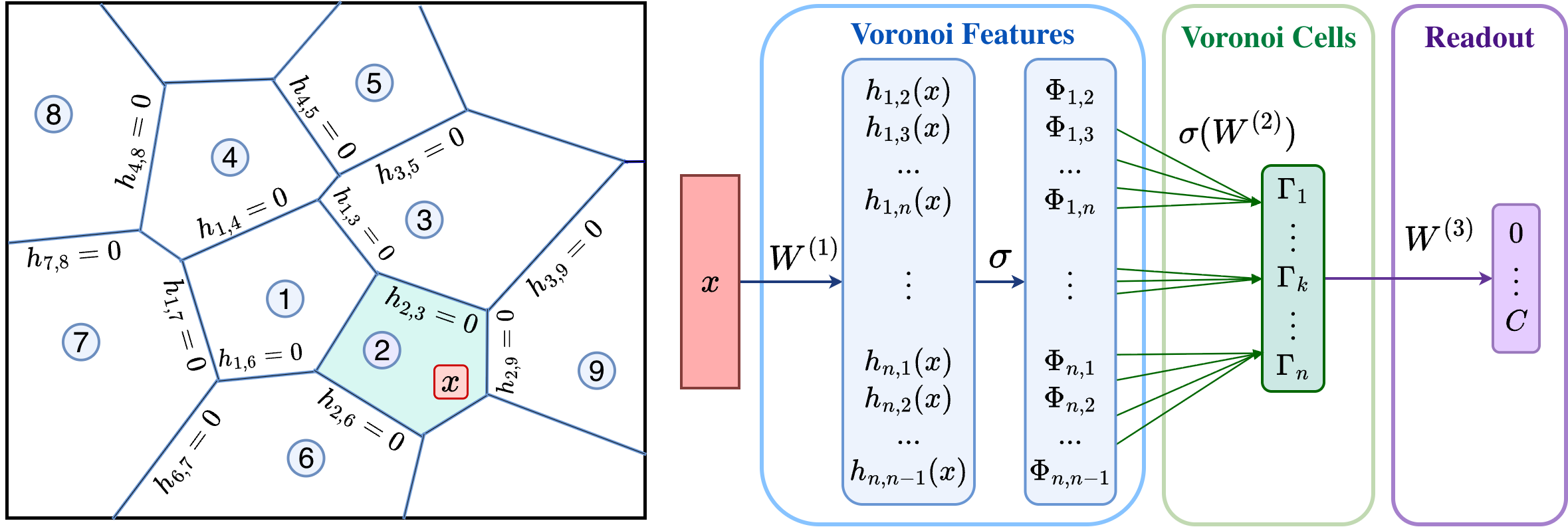}
    \caption{\textbf{Random Voronoi Network overview.}
    Pairwise bisectors $h_{ij}=0$ partition the dataset and define the Voronoi geometry. A query $x$ is mapped by $W^{(1)}$ and $\sigma$ to Voronoi features $\Phi_{ij}$, routed by $W^{(2)}$ into cell scores $\Gamma_i$ whose argmax identifies the nearest prototype, and mapped by $W^{(3)}$ to the final nearest-neighbor prediction.}
    \label{fig:method_diag}
\end{figure}

\section{Random Voronoi Networks}
\label{sec:rvn}

Random Voronoi Networks (RVNs) are explicit three-layer
multi-layer perceptrons whose weights are determined by Euclidean Voronoi
geometry. The first layer computes \emph{Voronoi features}, which are pairwise
comparisons between selected training samples. The second layer computes
\emph{Voronoi cells}, or more generally top-$k$ neighborhood indicators, by
aggregating those comparisons. The final layer averages the values attached to
the active cells. By changing only the threshold in the cell layer, the same
architecture recovers either $1$-nearest neighbor or $k$-nearest neighbors, assuming no distance ties.

\subsection{The \texorpdfstring{$k$}{k}-nearest-neighbor hypothesis class}
\label{subsec:knn-hypothesis-class}

Let $\mathcal{D}_N=\{(x_i,t_i)\}_{i=1}^N$ be a reference dataset of $N$ samples, with
query $x_i\in\R^d$ and target $t_i\in\R^p$. For regression, $t_i$ is a scalar or vector
response; for classification, $t_i$ may be a one-hot label vector. Write
$X=(x_1,\dots,x_N)$ and $T=(t_1,\dots,t_N)$. For a query $x\in\R^d$, let
$\mathcal{N}_N(x;X)=(i_1(x),\dots,i_N(x))$ be the ordering of the reference
samples by increasing squared Euclidean distance with respect to $X$, so that
$\|x-x_{i_1(x)}\|_2^2\le \cdots \le \|x-x_{i_N(x)}\|_2^2$. Ties are resolved
by a fixed deterministic rule. The $k$-nearest-neighbor set with respect to $X$
is
\begin{equation}
    \mathcal{N}_k(x;X)
    :=
    \{i_1(x),\dots,i_k(x)\}.
    \label{eq:k-neighborhood}
\end{equation}

The Euclidean $1$-NN predictor returns the value attached to the closest
reference sample:
\begin{equation}
    f_{1\mathrm{NN},\mathcal{D}_N}(x)
    :=
    t_{i_1(x)}
    =
    t_{\argmin_{i\in[N]}\|x-x_i\|_2^2},
    \label{eq:1nn-predictor}
\end{equation}
where the $\argmin$ uses the same deterministic tie-breaking rule. Thus $1$-NN
is a Voronoi-cell predictor: each input receives the value attached to the
sample whose Voronoi cell contains it.
The Euclidean $k$-NN predictor averages the values attached to the $k$ closest
reference samples:
\begin{equation}
    f_{k\mathrm{NN},\mathcal{D}_N}(x)
    :=
    \frac{1}{k}\sum_{i\in\mathcal{N}_k(x;X)} t_i .
    \label{eq:knn-predictor}
\end{equation}
For classification, when $t_i$ is one-hot,
$f_{k\mathrm{NN},\mathcal{D}_N}(x)$ stores empirical class frequencies among
the $k$ nearest neighbors, and prediction is
$\widehat y(x)=\argmax_{\ell\in[p]}[f_{k\mathrm{NN},\mathcal{D}_N}(x)]_\ell$.
We refer to the unrestricted Euclidean $k$-nearest-neighbor hypothesis class by 
$\mathcal{H}_{k\mathrm{NN}}^{d,p}$. For binary classification ($d\ge 1$, $k\ge 1$, with thresholded one-hot class scores $\mathcal{H}_{k\mathrm{NN},\mathrm{bin}}^{d}$), this class has infinite VC dimension \cite{shalev2014understanding}: for any nonempty finite set of query points, one can
place labeled reference sites arbitrarily close to each query point so as to
realize any binary labeling. Thus
$\mathrm{VCdim}(\mathcal{H}_{k\mathrm{NN},\mathrm{bin}}^{d})=\infty$.

\subsection{Random Voronoi Networks as three-layer MLPs}
\label{subsec:rvn-hypothesis-class}

RVNs are structured piece-wise parameterizations of nearest-neighbor geometry. The RVN is constructed using a set of anchors consisting of training samples $\{x_i:i\in A\}$, where
$A\subseteq [N]$ is a selected sample of indices with $|A|=M_C$. Let
$\mathcal{E}\subset A\times[N]$ be a set of ordered comparison pairs, with
$i\neq j$ for all $(i,j)\in\mathcal{E}$; the second index may be any reference
sample. Each pair $(i,j)$ defines one Voronoi
feature comparing sample $x_i$ against sample $x_j$. We write $M_F=|\mathcal{E}|$. The pairs in $\mathcal{E}$ are used to construct the Random Voronoi Feature layer $\Phi$, the set of anchors $A$ provide the basis for the Random Voronoi Cell layer $\Gamma$, and the set of labels $T$ are used to construct the final readout layer that completes the Random Voronoi Network $f_{\mathrm{RVN}}$.

The RVN can be written in the following structured network form, with
$\Phi:\R^d\to\R^{M_F}$, $\Gamma:\R^{M_F}\to\R^{M_C}$, and
$f:\R^{M_C}\to\R^p$.

In the step variant, $\Gamma(x)\in\{0,1\}^{M_C}$ is the vector of active cells. If $1\le k\le M_C$ and the comparison graph consists of all ordered pairs of distinct selected anchors, precisely $k$ cells are active whenever their distances to $x$ are pairwise distinct. For a fixed dataset, $\mathcal{H}_{\mathrm{RVN}}^{\sigma,k}(\mathcal{D}_N;M_C,M_F)$
denotes the family with fixed geometric hidden layers for each $(A,\mathcal E)$, activation $\sigma$, and free affine readout parameters.

\subsection{Random Voronoi Features}
\label{subsec:random-voronoi-features}

The first layer of an RVN constructs the Voronoi feature map
$\Phi(x)\in\R^{M_F}$. Its coordinates are indexed by the randomly selected ordered
comparison pairs $e=(i,j)\in\mathcal{E}$. Each coordinate compares the distance
from the query $x$ to a main anchor $x_i$ and a selected reference sample $x_j$. For each comparison pair, define the signed Voronoi margin:
\begin{equation}
\begin{aligned}
    h_{ij}(x)
    &:= \|x-x_j\|_2^2-\|x-x_i\|_2^2 \\
    &= 2(x_i-x_j)^\top x+\|x_j\|_2^2-\|x_i\|_2^2.
\end{aligned}
    \label{eq:random-voronoi-margin}
\end{equation}
Thus $h_{ij}(x)\ge 0$ exactly when $x_i$ is at least as close to $x$ as
$x_j$ is. Therefore, each Voronoi comparison is an affine neuron. For
$e=(i,j)$, the corresponding first-layer parameters are
\begin{equation}
    W^{(1)}_{e,:}=2(x_i-x_j)^\top,
    \qquad
    b^{(1)}_e=\|x_j\|_2^2-\|x_i\|_2^2,
    \label{eq:random-voronoi-feature-weights}
\end{equation}
so that $(W^{(1)}x+b^{(1)})_e=h_{ij}(x)$. The activation $\sigma$ determines how this comparison is encoded. With a
step activation $\sigma_H$,
\begin{equation}
\begin{aligned}
    \Phi^{\step}(x)
    &= \sigma_H(W^{(1)}x+b^{(1)}), \\
    \Phi_{ij}^{\step}(x)
    &= \sigma_H(h_{ij}(x)) \\
    &= \1\{\|x-x_i\|_2^2\le \|x-x_j\|_2^2\},
\end{aligned}
    \label{eq:random-voronoi-step-feature}
\end{equation}
so the feature records whether anchor $x_i$ is no farther than anchor $x_j$. With a ReLU
activation, the orientation is reversed so that the feature records the amount
by which the comparison for $x_i$ is violated:
\begin{equation}
\begin{aligned}
    \Phi^{\relu}(x)
    &= \relu\left(-(W^{(1)}x+b^{(1)})\right), \\
    \Phi_{ij}^{\relu}(x)
    &= \relu(-h_{ij}(x)).
\end{aligned}
    \label{eq:random-voronoi-relu-feature}
\end{equation}
Here the ReLU version uses the opposite orientation of the affine margin, so
that $\Phi_{ij}^{\relu}(x)$ is zero when $x_i$ is at least as close as $x_j$,
and positive otherwise. 
Thus the first layer stores the sampled Voronoi bisectors determined by
$\mathcal{E}$. The randomness selects which pairwise comparisons are included;
after $A$ and $\mathcal{E}$ are fixed, the weights and biases are completely
determined by the selected reference samples.

\subsection{Random Voronoi Cells}
\label{subsec:random-voronoi-cells}

The second layer aggregates pairwise Voronoi features into anchor-level cell
activations $\Gamma(x)\in\R^{M_C}$. For each selected anchor $x_i$, $i\in A$,
let $\mathcal{E}_i:=\{j\in[N]\setminus\{i\}:(i,j)\in\mathcal{E}\}$ denote its sampled rivals
and let $d_i:=|\mathcal{E}_i|$. Define the routing matrix
$W_C\in\{0,1\}^{M_C\times M_F}$ by
\begin{equation}
    (W_C)_{i,e}=\1\{e=(i,j)\text{ for some }j\in\mathcal{E}_i\}.
    \label{eq:random-voronoi-cell-routing}
\end{equation}
Thus $W_C$ sums exactly the comparisons associated with anchor $x_i$.

For the step RVN, choose $W^{(2)}=W_C$ and
$b^{(2)}_i=-d_i+k-\frac12$. Then
\begin{equation}
    \Gamma_i^{\step,k}(x)
    =
    \sigma_H\left(
        \sum_{j\in\mathcal{E}_i}\Phi_{ij}^{\step}(x)
        -d_i+k-\frac12
    \right).
    \label{eq:random-voronoi-cell-step}
\end{equation}
Hence $\Gamma_i^{\step,k}(x)=1$ exactly when $x_i$ loses to at most $k-1$ of
its sampled rivals, equivalently when it wins at least $d_i-k+1$ sampled
comparisons. If $\mathcal E=\mathcal E_A$ exactly, $1\le k\le M_C$, and
distances to the selected anchors are distinct, this activates precisely the
top-$k$ anchors within $A$; for $k=1$, it is their hard Voronoi-cell indicator.

For the ReLU RVN, the cell layer instead gates the accumulated comparison
violation. Since $\Phi_{ij}^{\relu}(x)=\relu(-h_{ij}(x))$ is zero when $x_i$ is
at least as close as $x_j$, define
\begin{equation}
    V_i(x):=(W_C\Phi^{\relu}(x))_i
    =
    \sum_{j\in\mathcal{E}_i}\Phi_{ij}^{\relu}(x).
    \label{eq:random-voronoi-accumulated-violation}
\end{equation}
Thus $V_i(x)=0$ when $x_i$ loses to none of its sampled rivals, and otherwise
measures the total violation margin. With gate width $\varepsilon>0$, choose
$W^{(2),\relu}_\varepsilon=-\varepsilon^{-1}W_C$ and
$b^{(2),\relu}_\varepsilon=\mathbf{1}_{M_C}$, giving
\begin{equation}
    \Gamma_i^{\relu,\varepsilon}(x)
    =
    \relu\left(1-\frac{V_i(x)}{\varepsilon}\right).
    \label{eq:random-voronoi-cell-relu-eps}
\end{equation}
This is a shrinking violation gate: it equals $1$ when the sampled comparisons
for $x_i$ are all satisfied, decays linearly for $0<V_i(x)<\varepsilon$, and
vanishes once $V_i(x)\ge\varepsilon$.
For fixed $\varepsilon>0$, this
ReLU map is differentiable
almost everywhere.

\subsection{Readout Layer}
\label{subsec:readout-layer}

The readout layer converts anchor-level cell activations into the final
prediction by attaching each selected anchor $x_i$, $i\in A$, to its target
$t_i\in\R^p$. For regression, $t_i=y_i$ is the response associated with
$x_i$; for classification, $t_i=e_{y_i}$ is the one-hot encoding of its class
label.

For the step RVN, $\Gamma^{\step,k}(x)$ is binary and is designed to activate
the selected neighbors. We therefore choose
$W^{(3),\step,k}_{:,i}=t_i/k$ and $b^{(3),\step,k}=0$, so that
\begin{equation}
    f^{\step,k}(x)
    =
    \frac{1}{k}\sum_{i\in A}\Gamma_i^{\step,k}(x)t_i.
    \label{eq:random-voronoi-readout-step}
\end{equation}
When exactly $k$ cells are active, this is the average of the targets attached to those anchors. However, under budgeted comparison sampling, the number of active cells may differ from $k$, so the readout may not be a normalized neighborhood average.

For the ReLU RVN, $\Gamma^{\relu,\varepsilon}(x)$ is generally soft and is
intended to select a single nearest anchor as $\varepsilon\to0^+$. We therefore
use the unnormalized readout $W^{(3),\relu}_{:,i}=t_i$ with
$b^{(3),\relu}=0$, giving
\begin{equation}
    f^{\relu,\varepsilon}(x)
    =
    \sum_{i\in A}\Gamma_i^{\relu,\varepsilon}(x)t_i.
    \label{eq:random-voronoi-readout-relu}
\end{equation}
For finite $\varepsilon$, this combines the targets of anchors with nonzero ReLU cell activity.

In classification, the output $f(x)\in\R^p$ is interpreted as a vector of class scores and prediction is $\widehat y(x)=\arg\max_c [f(x)]_c$. In regression, $f(x)$ is used directly as the predicted response.

\subsection{Equivalence between RVN and \texorpdfstring{$k$}{k}-NN}
\label{subsec:rvn-knn-equivalence}

We now state when RVNs exactly recover nearest-neighbor predictors.
Exact recovery uses the full-anchor, full-comparison endpoint, $M_C=N$ and
$M_F=N(N-1)$. For fixed finite $N$, this is a finite construction rather than
an asymptotic budget limit. The complete ordered graph is
\[
    \mathcal{E}_{\mathrm{full}}
    =
    \{(i,j):i,j\in[N],\ i\neq j\}.
\]
Let $\Omega_X\subset\R^d$ denote the set of query points with no distance ties
among the reference samples $X=(x_1,\ldots,x_N)$.

\begin{theorem}[Step RVNs recover $k$-NN]
\label{thm:step-rvn-knn-equivalence}
Fix a reference dataset
$\mathcal{D}_N=\{(x_i,t_i)\}_{i=1}^N$ and $1\le k\le N$. If $A=[N]$ and $\mathcal{E}=\{(i,j): i,j\in[N],\ i\neq j\},$ then the step RVN with the canonical readout $W^{(3)}_{:,i}=t_i/k$, $b^{(3)}=0,$ satisfies
\[
f_{\mathrm{RVN}}^{\step,k}(x)
=
f_{k\mathrm{NN},\mathcal{D}_N}(x)
\]
for every $x$ with no distance ties.
\end{theorem}

\begin{theorem}[ReLU RVNs recover $1$-NN at finite gate width]
\label{thm:relu-rvn-1nn-equivalence}
Fix a reference dataset $\mathcal{D}_N=\{(x_i,t_i)\}_{i=1}^N$
with $N\ge 1$. Suppose $A=[N]$ and
$\mathcal{E}=\mathcal{E}_{\mathrm{full}}$, and use the canonical
readout $W^{(3)}_{:,i}=t_i$, $b^{(3)}=0$.

Fix $x\in\Omega_X$. If $N\ge 2$, let $i_1(x)$ and $i_2(x)$
index the nearest and second-nearest reference samples, and define
their squared-distance gap by
$
    \Delta(x)
    :=
    \|x-x_{i_2(x)}\|_2^2-\|x-x_{i_1(x)}\|_2^2
    >0.
$
Then, for every $0<\varepsilon\le\Delta(x)$,
\begin{equation}
    f_{\mathrm{RVN}}^{\relu,\varepsilon}(x)
    =
    f_{1\mathrm{NN},\mathcal{D}_N}(x).
    \label{eq:relu-rvn-finite-gate-recovery}
\end{equation}
If $N=1$, the same equality holds for every $\varepsilon>0$.
Consequently, for every $x\in\Omega_X$,
\begin{equation}
    \lim_{\varepsilon\to0^+}
    f_{\mathrm{RVN}}^{\relu,\varepsilon}(x)
    =
    f_{1\mathrm{NN},\mathcal{D}_N}(x).
    \label{eq:relu-rvn-1nn-h-class-limit}
\end{equation}
\end{theorem}

The proofs are given in Appendix~\ref{app:proof-step-rvn-knn-limit} and
Appendix~\ref{app:proof-relu-rvn-1nn-limit}.

\subsection{Budgeted Approximation and Rank Geometry}
\label{subsec:randomness-and-complexity}

The anchor and comparison budgets control two distinct approximations. We write
$M_C=|A|$ for the number of main anchors and $M_F=M_C H$ when each main anchor
uses $H$ sampled rivals, with $1\le M_C\le N$ and
$0\le H\le N-1$. At the cell level, $H$ controls how accurately sampled
comparisons resolve $k$ nearest neighbors. At the feature level, we analyze how $M_F$ controls preservation of the complete pairwise ranking geometry under uniform random pair sampling.

\paragraph{Cell-level approximation.}
For each main anchor $i\in A$, $H$ rivals are sampled uniformly without replacement from $[N]\setminus\{i\}$. Define the global reference distance rank
\begin{equation}
    r_i^X(x)
    :=1+\sum_{j\in[N]\setminus\{i\}}
    \1\{\|x-x_j\|_2^2<\|x-x_i\|_2^2\},
\label{eq:rvn-reference-rank}
\end{equation}

so a reference sample has rank $1$ exactly when it attains the minimum distance to $x$. For a \emph{sampled} rival set
$\mathcal{E}_i\subseteq[N]\setminus\{i\}$ with $|\mathcal{E}_i|=H$, define

\begin{equation}
    L_i(x;\mathcal{E}_i)
    :=\sum_{j\in\mathcal{E}_i}
    \1\{\|x-x_j\|_2^2<\|x-x_i\|_2^2\}.
\label{eq:rvn-sampled-loss-count}
\end{equation}
Then, for $k\ge 1$, $\Gamma_i^{\step,k}(x)=\1\{L_i(x;\mathcal{E}_i)\le k-1\}$.
\begin{proposition}[Sampled comparisons control top-\(k\) activation]
\label{prop:rvn-hypergeom}
Fix $k\ge 1$, $0\le H\le N-1$, and a main anchor $i$ with reference set rank $r_i^X(x)$. If $\mathcal{E}_i\subseteq[N]\setminus\{i\}$ is sampled uniformly without replacement with $|\mathcal{E}_i|=H$, then
\begin{equation}
\Prob(\Gamma_i^{\step,k}(x)=1)=\Prob(L_i\le k-1), \text{ and}
\end{equation}
\begin{equation}
L_i(x;\mathcal{E}_i)
\sim
\operatorname{Hypergeom}(N-1,r_i^X(x)-1,H). 
\label{eq:rvn-hypergeom}
\end{equation}
\end{proposition}

Let $S_H^k(x):=\sum_{i\in A}\Gamma_i^{\step,k}(x)$ denote the number of active
main-anchor cells.

\begin{corollary}[Comparison budget controls effective neighborhood size]
\label{cor:rvn-effective-neighborhood}
Fix $k\ge 1$, $1\le M_C\le N$, and $0\le H\le N-1$. Suppose $A$ is sampled uniformly from $[N]$ with $|A|=M_C$, and each $i\in A$
samples $H$ rivals uniformly from $[N]\setminus\{i\}$. Assume that the reference samples have pairwise distinct distances to the query $x$. Then
\begin{equation}
\E[S_H^k(x)]
=M_C\frac{\min\{k,H+1\}}{H+1}.
\label{eq:rvn-effective-neighborhood-general}
\end{equation}
Practically, for $H\ge k-1$, define the effective neighborhood size
\begin{equation}
k_{\mathrm{eff}}
:=\E[S_H^k(x)]
=\frac{M_Ck}{H+1}
=\frac{M_C^2k}{M_F+M_C}.
\label{eq:rvn-effective-neighborhood}
\end{equation}
\end{corollary}

Together, Proposition~\ref{prop:rvn-hypergeom} and Corollary~\ref{cor:rvn-effective-neighborhood} quantify the effect of comparison subsampling: $H$ controls how reliably lower-ranked anchors are rejected, while $k_{\mathrm{eff}}=\E[S_H^k(x)]$ is the expected number of active cells and therefore the effective support of the approximate RVN. Smaller $H$ broadens this support, whereas $M_C=N$ and $H=N-1$ recover the exact $k$-NN support. This gives a direct, practical link between the feature budget and geometric smoothing; proofs are given in Appendix~\ref{app:budgeted-approximation-proofs}.

\paragraph{Feature-level approximation.}
The reference set $X=(x_1,\ldots,x_N)$ induces, for every no-tie input $x$, a
ranking $\pi_X(x)$ of all $N$ reference samples by increasing distance. For one
canonical orientation $i<j$, define
$\psi_{ij}(x)=2\Phi_{ij}^{\step}(x)-1\in\{-1,+1\}$ and let
$P=\binom{N}{2}$. A \emph{ranking disagreement} between two inputs is a pair
$\{i,j\}$ that the two induced rankings order differently.
\citet{pmlr-v37-jiao15} showed that normalized signed pairwise comparisons give an explicit feature map for the Kendall kernel. The following proposition identifies this map with complete signed step RVFs on reference-distance rankings.

\begin{proposition}[Complete RVFs encode Kendall rank geometry]
\label{prop:rvf-kendall-kernel}
Assume $N\ge2$ and $x,x'\in\Omega_X$. The complete signed step-RVF kernel satisfies
\begin{align}
K_X(x,x')
&:=\frac{1}{P}\sum_{i<j}\psi_{ij}(x)\psi_{ij}(x') \\
&=1-\frac{2D_K(\pi_X(x),\pi_X(x'))}{P},
\label{eq:rvf-kendall-kernel}
\end{align}
where $D_K$ counts pairwise ranking disagreements. Hence $K_X$ is the Kendall
rank correlation between the two reference-distance rankings and is positive
semidefinite.
\end{proposition}

ReLU RVFs retain the same binary ranking comparisons: for inputs without distance ties,
$\psi_{ij}(x)=1-2\cdot\mathbf{1}\{\Phi_{ij}^{\relu}(x)>0\}$.
Thus, the Kendall kernel is recoverable by thresholding ReLU coordinates, although it is not their raw inner product.

\begin{corollary}[Random RVFs preserve rank geometry with logarithmic feature growth]
\label{cor:rvf-kendall-concentration}
Assume $N\ge2$, and let $\mathcal Q=\{z_1,\ldots,z_m\}\subset\Omega_X$ be a fixed set of $m\ge1$ queries. 
Fix $\epsilon>0$, $0<\delta<1$, and an integer $M_F\ge1$. Sample $M_F$ canonical reference pairs independently and uniformly with replacement, and let
$\widehat K_{M_F}$ be their average signed comparison product. Then $\E[\widehat K_{M_F}(z_a,z_b)]=K_X(z_a,z_b)$ for all $a,b\in[m]$. With probability at least $1-\delta$,
\begin{equation}
\max_{a,b\in[m]}
|\widehat K_{M_F}(z_a,z_b)-K_X(z_a,z_b)|\le\epsilon
\label{eq:rvf-kendall-uniform-event}
\end{equation}
whenever
\begin{equation}
M_F\ge \frac{2}{\epsilon^2}\log\frac{2m^2}{\delta}.
\label{eq:rvf-kendall-finite-sample}
\end{equation}
\end{corollary}

Proposition~\ref{prop:rvf-kendall-kernel} and 
Corollary~\ref{cor:rvf-kendall-concentration} show that complete step RVFs encode the full reference-distance ranking through a Kendall kernel, while $M_F=O\!\left(\epsilon^{-2}[\log m+\log(1/\delta)]\right)$ uniformly preserves these similarities over $m$ fixed queries. Thus, under uniform independent pair sampling, a finite random RVF map can approximate ranking geometry without materializing all $\binom{N}{2}$ canonical comparisons. This provides a baseline theoretical justification for comparison subsampling, while the structured selectors considered below depart from this sampling model and are evaluated empirically. Proofs are given in Appendix~\ref{app:budgeted-approximation-proofs}.

\begin{figure}[t]
    \centering
    \includegraphics[width=\linewidth]{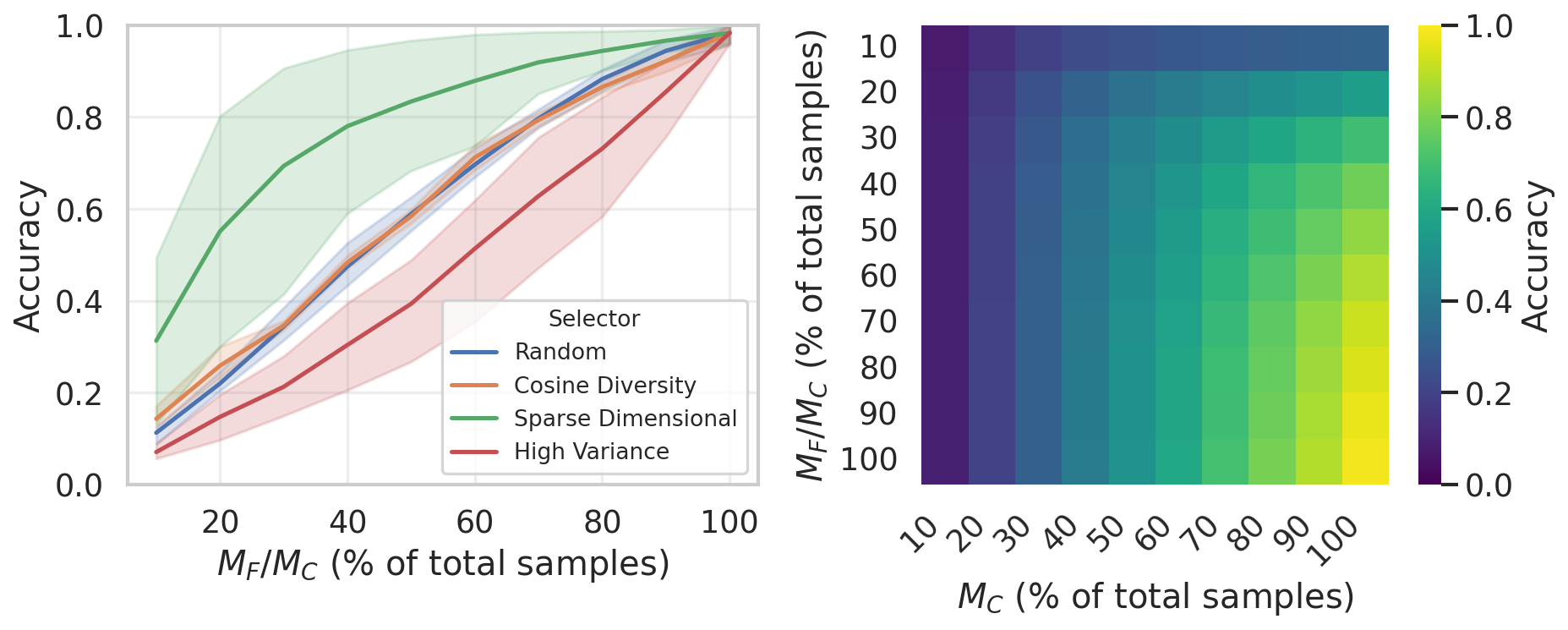}
    \caption{\textbf{RVNs recover exact nearest-neighbor indices.}
    \textit{Left:} index-matching accuracy versus $H=M_F/M_C$; Sparse Dimension performs best at small budgets, while all selectors converge to exact agreement.
    \textit{Right:} accuracy versus $M_C$ and $H$, reaching $1$ as both approach full coverage.}
    \label{fig:1nn}
\end{figure}

\subsection{Feature-pair selectors for approximate RVF}
\label{subsec:rvf-selectors}

For each main anchor $i\in A$, a selector chooses $H$ distinct rivals $S_i\subset[N]\setminus\{i\}$ and forms
\(
    \mathcal{E}
    =
    \bigcup_{i\in A}\{(i,j):j\in S_i\}.
\)
The selectors below differ only in how these $H$ rivals are chosen. All are unsupervised and label-agnostic, and aim to use the fixed comparison budget efficiently.

\textbf{Cosine Diversity selector.}
For a main anchor $i$ and candidate $j$, let $u_{ij}=(x_j-x_i)/\|x_j-x_i\|_2$. The Cosine Diversity selector greedily accepts $j$ when
\begin{equation}
    \max_{\ell\in S_i}|\langle u_{ij},u_{i\ell}\rangle|<\tau,
    \qquad
    |S_i|<M_F/M_C,
    \label{eq:diverse-selector-condition}
\end{equation}
for $\tau\in(0,1)$, filling any remaining slots uniformly at random. This discourages nearly parallel comparisons and spreads each cell across more angular directions.

\textbf{Sparse Dimension selector.}
The Sparse Dimension selector favors pairs whose differences are concentrated in few coordinates. For $d_{ij}=|x_i-x_j|\in\R^d_{\ge0}$, score candidates by Hoyer sparsity~\citep{hoyer2004non},
\begin{equation}
    \operatorname{sp}(d_{ij})
    =
    \frac{\sqrt{d}-\|d_{ij}\|_1/(\|d_{ij}\|_2+\epsilon)}{\sqrt{d}-1},
    \qquad
    \epsilon>0
    \label{eq:hoyer-sparsity}
\end{equation}
with the following conventions: set $\operatorname{sp}(0)=1$ in every dimension. When $d=1$ and $d_{ij}\neq 0$, set $\operatorname{sp}(d_{ij})=0$. For each $i$, the selector chooses the $M_F/M_C$ candidates with largest $\operatorname{sp}(d_{ij})$. These comparisons produce more axis-aligned and interpretable Voronoi hyperplanes.

\textbf{High Variance selector.}
The High Variance selector prioritizes comparisons whose bisectors split the empirical data mass. For a candidate pair $(i,j)$, define $\chi_{ij}(z)=\1\{\|z-x_i\|_2^2\le\|z-x_j\|_2^2\}$. On a subset $\mathcal{Z}\subseteq\mathcal{D}$, compute
\begin{equation}
    \widehat p_{ij}
    =
    |\mathcal{Z}|^{-1}\sum_{z\in\mathcal{Z}}\chi_{ij}(z),
    \qquad
    q_{ij}=\widehat p_{ij}(1-\widehat p_{ij}).
    \label{eq:variance-selector-score}
\end{equation}
For each $i$, the selector chooses the $M_F/M_C$ candidates with largest $q_{ij}$. Since $q_{ij}$ is maximized near $\widehat p_{ij}=1/2$, selected comparisons tend to cut through the data rather than produce nearly constant activations.
The four selectors trade off simplicity, angular coverage, coordinate sparsity, and empirical informativeness under the same budgets $M_C$ and $M_F$. Detailed implementations are given in Appendix~\ref{appendix:selectors}.

\section{Experimental Results}
\label{sec:experiments}

We evaluate three aspects of RVNs: nearest-neighbor recovery, RVF as a fixed tabular encoding, and geometric interpretability. Experimental details are given in Appendix~\ref{appendix:rvf-supervised-details}, with compute resources in Appendix~\ref{appendix:compute-resources}.

\begin{figure}[t]
    \centering
    \includegraphics[width=\linewidth]{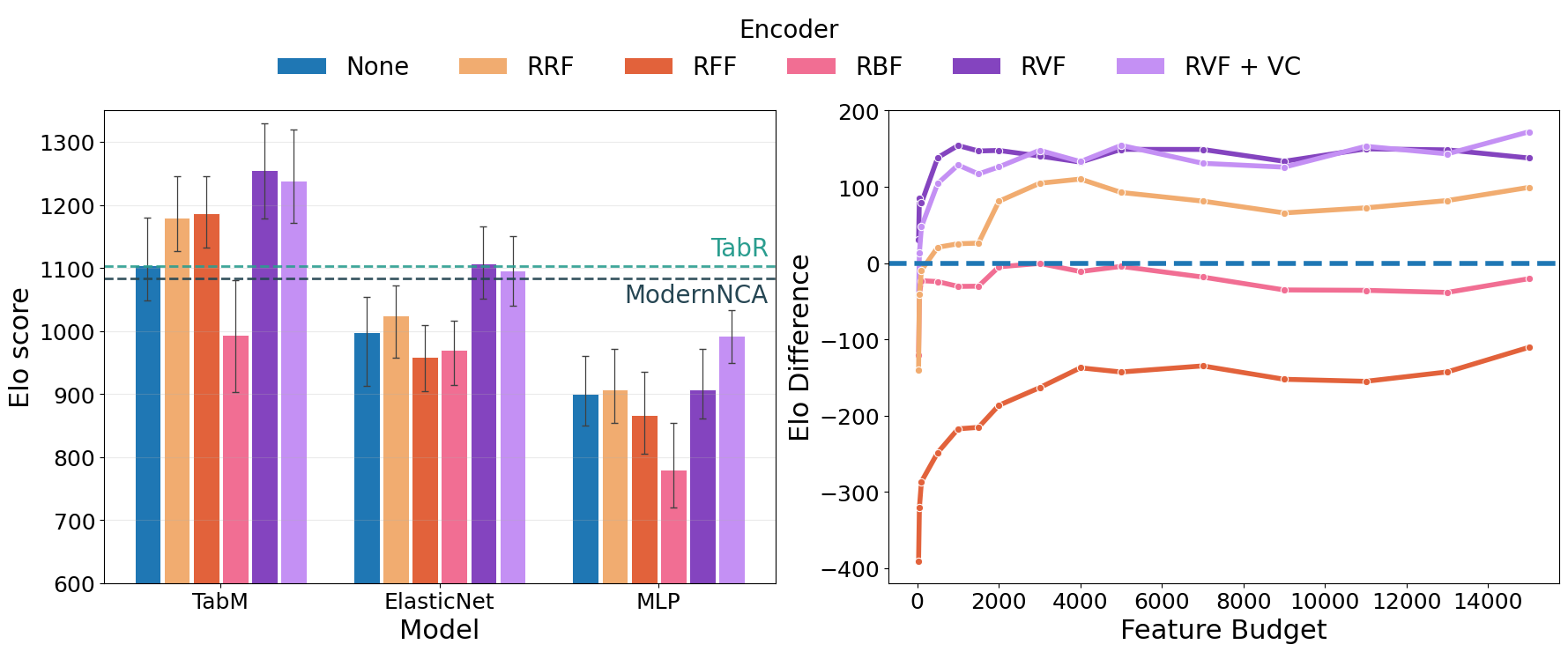}
    \caption{\textbf{RVFs improve downstream tabular performance.}
    \textit{Left:} aggregate Elo across downstream models show that RVF encodings outperform benchmark feature maps.
    \textit{Right:} ElasticNet Elo versus feature count compared to raw baseline, showing an early performance elbow and efficient data-dependent encoding.}
    \label{fig:elo}
\end{figure}

\subsection{Equivalence between RVN and 1-NN}
\label{subsec:rvf-1nn-comparison}

We verify that RVF construction selects the same nearest-neighbor index as exact Euclidean $1$-NN. For each query $x$, we compare
\(
    i_{\mathrm{1NN}}(x)
    =
    \argmin_{i\in[N]}\|x-x_i\|_2^2,
   \) and \(
    i_{\mathrm{RVF}}(x)
    =
    \argmax_{i\in[N]}\Gamma_i(x),
\) with 
\(
    \mathrm{Acc}
    =
    \frac{1}{|\mathcal{Q}|}
    \sum_{x\in\mathcal{Q}}
    \1\{i_{\mathrm{RVF}}(x)=i_{\mathrm{1NN}}(x)\},
    \label{eq:rvf-1nn-index-matching}
\)
where $\Gamma_i(x)$ is the Voronoi-cell score for sample $i$. 

Consistent with the cell-level approximation theory in Section~\ref{subsec:randomness-and-complexity}, Figure~\ref{fig:1nn} shows that increasing the rival budget $H=M_F/M_C$ improves rank resolution by reducing the chance that a lower-ranked anchor misses closer competitors, while increasing $M_C$ increases anchor coverage. Sparse Dimension selection gives substantially better recovery at small rival budgets, showing that structured allocation can identify useful comparisons more efficiently than uniform sampling. All selectors converge to exact $1$-NN when both anchor and comparison coverage become complete.

\subsection{Random Voronoi Feature Encoding}
\label{subsec:rvf-supervised-learning}
We next evaluate RVF as a fixed, label-agnostic, data-dependent encoding for supervised tabular learning. Given an encoder $\mathcal{T}:\R^d\to\R^q$, we train downstream predictors on transformed inputs $\mathcal{T}(x)$. We compare raw inputs, Random Fourier Features (RFF)~\citep{rahimi2007random}, Random ReLU Features (RRF)~\citep{hsu2021random}, Nyström RBF Features (RBF), and two RVF variants:
\begin{equation}
\begin{aligned}
    \mathcal{T}_{\mathrm{RVF}}(x)
    &=\Phi^{\relu}(x), \\
    \mathcal{T}_{\mathrm{RVF+VC}}(x)
    &=\bigl[\Phi^{\relu}(x);\Gamma^{\relu, \varepsilon}(x)\bigr].
\end{aligned}
    \label{eq:rvf-supervised-encoders}
\end{equation}
RVF uses the pairwise ReLU margins $\Phi^{\relu}$, while RVF+VC additionally includes the aggregated cell representation $\Gamma^{\relu, \varepsilon}$. We evaluate on TabArena-Lite~\citep{erickson2025tabarena} with ElasticNet, TorchMLP, and TabM-Mini~\citep{gorishniy2025tabm}; TabR~\citep{gorishniy2024tabr} and ModernNCA~\citep{ye2025modernnca} are included as specialized learned neighborhood-model references. Full protocol details are in Appendix~\ref{appendix:rvf-supervised-details}.

Figure~\ref{fig:elo} shows that RVF-based encoders achieve the strongest aggregate rankings among the evaluated feature maps. ElasticNet benefits strongly, indicating that RVF makes useful local structure more linearly accessible. 
Interestingly, ModernNCA and TabR, which employ learned neighborhood representations, achieve performance comparable to ElasticNet with RVF. We view this comparison as indication that RVF can inject a strong local geometric bias into non-specialized models. 

The feature-budget curve further shows a strong performance elbow with comparatively few encoded features. Overall, these results support the role of RVF as a reusable representation rather than merely a neural reproduction of Euclidean $k$-NN. A classifier-versus-encoder ablation is reported in Appendix~\ref{appendix:classifier-encoder-ablation}, and individual dataset results are shown in Appendix~\ref{appendix:individual_datasets}.

\begin{figure}[t]
    \centering
    \includegraphics[width=\linewidth]{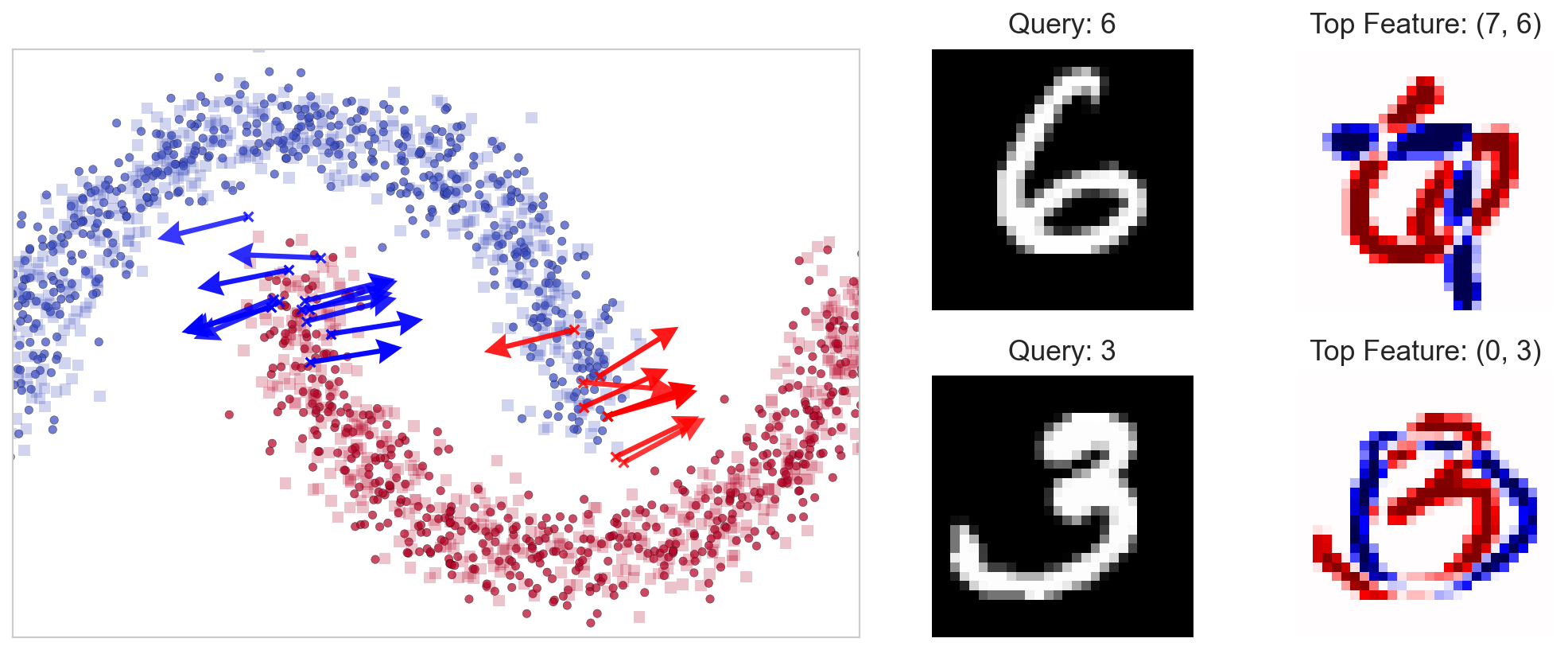}
    \caption{\textbf{RVF interpretability.}
    \textit{Left:} top $15$ logistic-regression RVFs on moons, shown as directions $v_{ij}=x_j-x_i$ at midpoints $m_{ij}$; influential features concentrate near discriminative regions and align with class separation.
    \textit{Right:} MNIST examples showing the highest-contributing RVF for each query.}
    \label{fig:rvf-interpret}
\end{figure}

\subsection{Interpretability of Random Voronoi Features}
\label{subsec:rvf-interpretability}

RVFs support interpretation at both global and query-specific levels,
since each coordinate corresponds to a reference pair $(x_i,x_j)$,
with comparison direction $v_{ij}=x_j-x_i$ and bisector midpoint
$m_{ij}=(x_i+x_j)/2$.
For a linear classifier with class scores
$
s_c(x)=b_c+\sum_e \beta_{c,e}z_e(x),
$
where $z_e(x)$ denotes the encoded feature supplied to the classifier,
coefficient magnitudes $|\beta_{c,e}|$ identify globally influential
comparisons, while
$a_{c,e}(x)=\beta_{c,e}z_e(x)$ gives each comparison's signed
contribution to the score for a particular query.

Figure~\ref{fig:rvf-interpret} illustrates both levels.
On moons, we display the comparisons with the largest coefficient magnitudes, revealing global directions concentrated near discriminative regions.
On MNIST, a simple logistic regression model trained on RVF features reaches $91\%$ accuracy. We select the feature with the largest absolute contribution $|a_{\hat c,e}(x)|$ to the predicted class score for each displayed query, and visualize its reference pair and signed image difference.
The strongest feature often contains an anchor from the same class as the query, providing a visually meaningful local comparison. Vision-specific modeling is outside our scope, but this experiment illustrates that RVF-based encodings are inherently inspectable: predictions can be decomposed into sample-pair comparisons and local directions.

\section{Limitations and Conclusion}
\label{sec:conclusion}
\label{sec:discussion-limitations}

RVNs benefit from locality, making them useful for piecewise or prototype-like tasks and naturally interpretable through sample-pair comparisons. However, they can suffer in high dimensions and scale poorly if all pairwise anchor comparisons are materialized. Practical variants should use better representations or metrics, subsampling, pruning, distance-based computation, or approximate comparisons. Future extensions include learned Mahalanobis metrics, anisotropic cells, and task-adaptive selectors to better align Voronoi partitions with real tabular data geometry.

Random Voronoi Networks provide a structured neural formulation of nearest-neighbor geometry through pairwise affine comparisons and Voronoi-cell aggregation. Step RVNs recover Euclidean $k$-NN at the full-comparison endpoint, while ReLU RVNs provide a differentiable $1$-NN relaxation. Budgeted RVFs retain this geometry with far fewer comparisons and improve downstream tabular models while remaining interpretable. Overall, RVNs connect nearest-neighbor methods, random features, and neural networks through a common Voronoi construction.

\clearpage
\section*{AI Use Statement}
AI tools were used to assist with drafting, editing, and improving the presentation of this manuscript, including refinement of prose, organization, and mathematical formulation. All content, claims, derivations, and results were reviewed and verified by the authors who take full responsibility for the accuracy and content of the paper.

\bibliographystyle{plainnat}
\bibliography{refs}

@inproceedings{baras1990convergence,
  author = {Baras, John S. and LaVigna, Anthony},
  title = {Convergence of a neural network classifier},
  booktitle = {Advances in Neural Information Processing Systems 3},
  pages = {839--845},
  year = {1990}
}

@article{cover1967nearest,
  author = {Cover, Thomas M. and Hart, Peter E.},
  title = {Nearest neighbor pattern classification},
  journal = {IEEE Transactions on Information Theory},
  volume = {13},
  number = {1},
  pages = {21--27},
  year = {1967}
}

@book{devroye1996probabilistic,
  author = {Devroye, Luc and Gy\"orfi, L\'aszl\'o and Lugosi, G\'abor},
  title = {A Probabilistic Theory of Pattern Recognition},
  publisher = {Springer},
  year = {1996}
}

@inproceedings{gorishniy2024tabr,
  author = {Gorishniy, Yury and Rubachev, Ivan and Kartashev, Nikolay and Shlenskii, Daniil and Kotelnikov, Akim and Babenko, Artem},
  title = {TabR: Tabular deep learning meets nearest neighbors},
  booktitle = {International Conference on Learning Representations},
  year = {2024}
}

@inproceedings{hsu2021random,
  author = {Hsu, Daniel and Sanford, Clayton H. and Servedio, Rocco and Vlatakis-Gkaragkounis, Emmanouil V.},
  title = {On the approximation power of two-layer networks of random ReLUs},
  booktitle = {Proceedings of the Thirty Fourth Conference on Learning Theory},
  series = {Proceedings of Machine Learning Research},
  volume = {134},
  pages = {2423--2461},
  year = {2021}
}

@book{okabe1992spatial,
  author = {Okabe, Atsuyuki and Boots, Barry and Sugihara, Kokichi},
  title = {Spatial Tessellations: Concepts and Applications of Voronoi Diagrams},
  publisher = {John Wiley \& Sons},
  year = {1992}
}

@article{plotz2018n3,
  author = {Pl\"otz, Tobias and Roth, Stefan},
  title = {Neural nearest neighbors networks},
  journal = {arXiv preprint arXiv:1810.12575},
  year = {2018}
}

@inproceedings{rahimi2007random,
  author = {Rahimi, Ali and Recht, Benjamin},
  title = {Random features for large-scale kernel machines},
  booktitle = {Advances in Neural Information Processing Systems 20},
  year = {2007}
}

@article{saralajew2019pbnn,
  author = {Saralajew, Sascha and Holdijk, Lars and Rees, Maike and Villmann, Thomas},
  title = {Prototype-based neural network layers: Incorporating vector quantization},
  journal = {arXiv preprint arXiv:1812.01214},
  year = {2019}
}

@inproceedings{sato1996glvq,
  author = {Sato, Atsushi and Yamada, Keiji},
  title = {Generalized learning vector quantization},
  booktitle = {Advances in Neural Information Processing Systems 8},
  pages = {423--429},
  year = {1996}
}

@inproceedings{sitawarin2021voronoi,
  author = {Sitawarin, Chawin and Kornaropoulos, Evgenios M. and Song, Dawn and Wagner, David},
  title = {Adversarial examples for k-nearest neighbor classifiers based on higher-order Voronoi diagrams},
  booktitle = {Advances in Neural Information Processing Systems 34},
  year = {2021}
}

@article{stone1977consistent,
  author = {Stone, Charles J.},
  title = {Consistent nonparametric regression},
  journal = {The Annals of Statistics},
  volume = {5},
  number = {4},
  pages = {595--620},
  year = {1977}
}

@inproceedings{voracek2022robust,
  author = {Vor\'a\v{c}ek, V\'aclav and Hein, Matthias},
  title = {Provably adversarially robust nearest prototype classifiers},
  booktitle = {Proceedings of the 39th International Conference on Machine Learning},
  series = {Proceedings of Machine Learning Research},
  volume = {162},
  pages = {22019--22057},
  year = {2022}
}

@article{hoyer2004non,
  author  = {Patrik O. Hoyer},
  title   = {Non-negative Matrix Factorization with Sparseness Constraints},
  journal = {Journal of Machine Learning Research},
  volume  = {5},
  pages   = {1457--1469},
  year    = {2004}
}

@inproceedings{erickson2025tabarena,
  author    = {Nick Erickson and Lennart Purucker and Andrej Tschalzev and
               David Holzmüller and Prateek Desai and David Salinas and
               Frank Hutter},
  title     = {TabArena: A Living Benchmark for Machine Learning on Tabular Data},
  booktitle = {Advances in Neural Information Processing Systems},
  volume    = {38},
  year      = {2025},
  doi       = {10.52202/085713-0519}
}

@book{elo1966uscf,
  title={The USCF Rating System: Its Development, Theory, and Applications},
  author={Elo, A.E.},
  url={https://books.google.es/books?id=onUazQEACAAJ},
  year={1966},
  publisher={United States Chess Federation}
}

@inproceedings{akiba2019optuna,
  title={{O}ptuna: A Next-Generation Hyperparameter Optimization Framework},
  author={Akiba, Takuya and Sano, Shotaro and Yanase, Toshihiko and Ohta, Takeru and Koyama, Masanori},
  booktitle={The 25th ACM SIGKDD International Conference on Knowledge Discovery \& Data Mining},
  pages={2623--2631},
  year={2019}
}

@book{shalev2014understanding,
  title={Understanding machine learning: From theory to algorithms},
  author={Shalev-Shwartz, Shai and Ben-David, Shai},
  year={2014},
  publisher={Cambridge university press}
}

@article{graepel2021pacbayesian,
  title        = {A PAC-Bayesian Analysis of Distance-Based Classifiers: Why Nearest-Neighbour Works!},
  author       = {Graepel, Thore and Herbrich, Ralf},
  journal      = {arXiv preprint arXiv:2109.13889},
  year         = {2021},
  doi          = {10.48550/arXiv.2109.13889},
  url          = {https://arxiv.org/abs/2109.13889}
}

@inproceedings{chaudhuri2014rates,
  title        = {Rates of Convergence for Nearest Neighbor Classification},
  author       = {Chaudhuri, Kamalika and Dasgupta, Sanjoy},
  booktitle    = {Advances in Neural Information Processing Systems},
  volume       = {27},
  pages        = {3437--3445},
  year         = {2014},
  publisher    = {Curran Associates, Inc.}
}

@article{murphy1990nearest,
  author  = {Murphy, Owen J.},
  title   = {Nearest Neighbor Pattern Classification Perceptrons},
  journal = {Proceedings of the IEEE},
  volume  = {78},
  number  = {10},
  pages   = {1595--1598},
  year    = {1990}
}

@article{bose1993neural,
  author  = {Bose, Nirmal K. and Garga, Amulya K.},
  title   = {Neural Network Design Using Voronoi Diagrams},
  journal = {IEEE Transactions on Neural Networks},
  volume  = {4},
  number  = {5},
  pages   = {778--787},
  year    = {1993}
}

@article{chen1997neural,
  author  = {Chen, Yan Qiu and Damper, Robert I. and Nixon, Mark S.},
  title   = {On Neural-Network Implementations of k-Nearest Neighbor Pattern Classifiers},
  journal = {IEEE Transactions on Circuits and Systems I: Fundamental Theory and Applications},
  volume  = {44},
  number  = {7},
  pages   = {622--629},
  year    = {1997}
}

@article{gentile2001improved,
  author  = {Gentile, Camillo and Sznaier, Mario},
  title   = {An Improved Voronoi-Diagram-Based Neural Net for Pattern Classification},
  journal = {IEEE Transactions on Neural Networks},
  volume  = {12},
  number  = {5},
  pages   = {1227--1234},
  year    = {2001}
}

@inproceedings{gentile2002hyperborders,
  author    = {Gentile, Camillo and Sznaier, Mario},
  title     = {Hyperborders in the Voronoi-Diagram-Based Neural Net for Pattern Classification},
  booktitle = {Proceedings of the 2002 International Joint Conference on Neural Networks (IJCNN'02)},
  volume    = {3},
  publisher = {IEEE},
  year      = {2002}
}

@inproceedings{ye2025modernnca,
  author    = {Ye, Han-Jia and others},
  title     = {Revisiting Nearest Neighbor for Tabular Data: A Deep Tabular Baseline Two Decades Later},
  booktitle = {International Conference on Learning Representations},
  year      = {2025}
}

@inproceedings{williamsnystromrbf,
  author    = {Christopher K. I. Williams and Matthias Seeger},
  title     = {Using the Nyström Method to Speed Up Kernel Machines},
  booktitle = {Advances in Neural Information Processing Systems},
  volume    = {13},
  pages     = {682--688},
  year      = {2001}
}

@inproceedings{
gorishniy2025tabm,
title={TabM: Advancing tabular deep learning with parameter-efficient ensembling},
author={Yury Gorishniy and Akim Kotelnikov and Artem Babenko},
booktitle={The Thirteenth International Conference on Learning Representations},
year={2025},
url={https://openreview.net/forum?id=Sd4wYYOhmY}
}

@inproceedings{pmlr-v37-jiao15,
  title     = {The Kendall and Mallows Kernels for Permutations},
  author    = {Jiao, Yunlong and Vert, Jean-Philippe},
  booktitle = {Proceedings of the 32nd International Conference on Machine Learning},
  editor    = {Bach, Francis and Blei, David},
  series    = {Proceedings of Machine Learning Research},
  volume    = {37},
  pages     = {1935--1944},
  year      = {2015},
  publisher = {PMLR},
  url       = {https://proceedings.mlr.press/v37/jiao15.html}
}

@inproceedings{10.1007/978-3-642-03784-9_34,
  author    = {Bachrach, Yoram and Herbrich, Ralf and Porat, Ely},
  editor    = {Karlgren, Jussi and Tarhio, Jorma and Hyyr{\"o}, Heikki},
  title     = {Sketching Algorithms for Approximating Rank Correlations in Collaborative Filtering Systems},
  booktitle = {String Processing and Information Retrieval},
  publisher = {Springer Berlin Heidelberg},
  address   = {Berlin, Heidelberg},
  pages     = {344--352},
  year      = {2009},
  doi       = {10.1007/978-3-642-03784-9_34}
}

\clearpage
\appendix

\section{RVN Formulation}

\subsection{The \texorpdfstring{$k$}{k}-nearest-neighbor hypothesis class}
\label{appendix:knn-hypothesis-class}

Let $\mathcal{D}_N=\{(x_i,t_i)\}_{i=1}^N$ be a reference dataset with
$x_i\in\R^d$ and $t_i\in\R^p$, and write $X=(x_1,\dots,x_N)$ and
$T=(t_1,\dots,t_N)$. For a query $x\in\R^d$, let
$(i_1(x),\dots,i_N(x))$ be the deterministic ordering of the reference points
by increasing squared Euclidean distance to $x$, breaking distance ties by increasing dataset index, and define
$\mathcal{N}_k(x;X):=\{i_1(x),\dots,i_k(x)\}$. The Euclidean $k$-NN predictor is
\begin{equation}
    f_{k\mathrm{NN},\mathcal{D}_N}(x)
    :=
    \frac{1}{k}\sum_{i\in\mathcal{N}_k(x;X)} t_i .
    \label{eq:app-knn-predictor}
\end{equation}
For regression this is the average response among the $k$ nearest reference
points. For classification with one-hot labels, it gives empirical class
frequencies and predicts
$\widehat y(x)=\argmax_{\ell\in[p]}[f_{k\mathrm{NN},\mathcal{D}_N}(x)]_\ell$.
The case $k=1$ is the usual Voronoi-cell nearest-neighbor rule.

For a fixed dataset $\mathcal{D}_N$, the expression above defines one
predictor. The corresponding $N$-site Euclidean $k$-nearest-neighbor hypothesis
class is obtained by allowing both the reference sites and their attached
values to vary:
\begin{equation}
\begin{aligned}
    \mathcal{H}_{k\mathrm{NN}}^{N,d,p}
    := \Bigl\{&x\mapsto \frac{1}{k}\sum_{i\in\mathcal{N}_k(x;X)} t_i :\\
        &X\in(\R^d)^N,\; T\in(\R^p)^N\Bigr\}.
\end{aligned}
    \label{eq:knn-hypothesis-class-fixed-N}
\end{equation}

The unrestricted, nonparametric $k$-NN class allows the number of reference
sites to vary:
\begin{equation}
    \mathcal{H}_{k\mathrm{NN}}^{d,p}
    :=
    \bigcup_{N\ge k}\mathcal{H}_{k\mathrm{NN}}^{N,d,p}.
    \label{eq:knn-hypothesis-class}
\end{equation}
For binary classification with $d\ge 1$ and $k\ge 1$ (denoting the resulting binary classifier class by $\mathcal{H}_{k\mathrm{NN},\mathrm{bin}}^{d}$), this unrestricted class has infinite VC dimension.
Indeed, given any $m\ge 1$ distinct points $z_1,\dots,z_m\in\R^d$ and any binary
labeling, place $k$ reference sites sufficiently close to each $z_j$, all with
the prescribed label of $z_j$, choosing the neighborhoods disjoint. Then the
$k$ nearest neighbors of each $z_j$ are exactly its local reference sites, so
the classifier realizes the prescribed labeling. Since this shatters finite
sets of arbitrary size,
\begin{equation}
    \mathrm{VCdim}\!\left(
        \mathcal{H}_{k\mathrm{NN},\mathrm{bin}}^{d}
    \right)
    =
    \infty .
    \label{eq:knn-infinite-vc-dim}
\end{equation}
This standard shattering argument is often cited as the reason that
unrestricted nearest-neighbor classifiers have infinite VC dimension; see, for
example, \cite{shalev2014understanding}.

\subsection{Three-layer MLPs RVN Hypothesis Class}
\label{app:rvn-hypothesis-class}

This appendix subsection makes explicit the hypothesis class induced by the
three-layer Random Voronoi Network (RVN) construction. The purpose is to separate
three sources of variation: the choice of anchors, the choice of sampled
Voronoi comparisons, and the free readout parameters. Throughout, the reference
dataset
\[
    \mathcal{D}_N=\{(x_i,t_i)\}_{i=1}^N
\]
is fixed. Let $A \subseteq[N]$ denote the selected anchor indices, with
$|A|=M_C$, and let
\[
    \mathcal{E}\subseteq \{(i,j):i\in A,\ j\in[N]\setminus\{i\}\},
    \qquad
    |\mathcal{E}|=M_F,
\]
denote the selected ordered comparison graph. The hidden-layer parameters of
the RVN are determined by $(A,\mathcal{E})$ and the Euclidean geometry of the
anchor points. The only trainable, or freely chosen, parameters in the class
below are the readout weights and biases.

For an activation choice $\sigma$ and neighborhood parameter $k$, define
\begin{equation}
\begin{aligned}
    &\mathcal{H}_{\mathrm{RVN}}^{\sigma,k}(\mathcal{D}_N;M_C,M_F)\\
    &:= \Bigl\{f_{A,\mathcal{E},\Theta}^{\sigma,k}:\\
    &\qquad A \subseteq[N],\ |A|=M_C,\\
    &\qquad \mathcal{E}\subset A\times[N],\ |\mathcal{E}|=M_F,\\
    &\qquad i\neq j\text{ for all }(i,j)\in\mathcal{E},\\
    &\qquad \Theta=(W^{(3)},b^{(3)})\in\R^{p\times M_C}\times\R^p
    \Bigr\}.
\end{aligned}
\label{eq:rvn-hypothesis-class}
\end{equation}
Each function in the class has the three-layer form
\begin{equation}
    f_{A,\mathcal{E},\Theta}^{\sigma,k}(x)
    =
    W^{(3)}\Gamma_{A,\mathcal{E}}^{\sigma,k}(x)+b^{(3)},
    \label{eq:app-rvn-three-layer-form}
\end{equation}
where the feature layer computes sampled Voronoi comparisons and the cell layer
aggregates those comparisons into anchor-level activations. Thus the class is a
structured subfamily of three-layer MLPs: it has the same input--hidden--hidden--
output architecture as a standard MLP, but the first two layers are not arbitrary
parameters. They are fixed by the selected anchors and by the selected
comparison graph.

The quantity $M_C$ controls the number of prototype cells available to the
network, while $M_F$ controls the number of pairwise Voronoi constraints used to
define those cells. In the full-anchor and full-comparison case,
\[
    A=[N],
    \qquad
    \mathcal{E}_{full}
    =
    \{(i,j)\in[N]\times[N]:i\neq j\},
\]
the step version recovers the usual Euclidean $k$-nearest-neighbor rule, up to
the deterministic tie convention. For strict statements below, we work away from
distance ties, or equivalently assume a fixed perturbation or tie-breaking rule.

\subsubsection{Step RVN}
\label{subsec:step-rvn}

The step RVN is the hard-cell version of the construction. It uses the
Heaviside activation
\[
    H(u)=\1\{u\ge 0\}
\]
in both hidden layers. For each ordered comparison pair $(i,j)\in\mathcal{E}$,
the first layer computes the binary Voronoi comparison
\begin{equation}
\begin{aligned}
    \Phi_{ij}^{\step}(x)
    &= H\!\left(\|x-x_j\|_2^2-\|x-x_i\|_2^2\right) \\
    &= \1\{\|x-x_i\|_2^2\le \|x-x_j\|_2^2\}.
\end{aligned}
    \label{eq:app-step-rvn-feature}
\end{equation}
Thus $\Phi_{ij}^{\step}(x)=1$ exactly when anchor $x_i$ is no farther than anchor $x_j$ in
distance to the query.

For each anchor $i\in A$, write
\[
    \mathcal{E}_i:=\{j\in[N]\setminus\{i\}:(i,j)\in\mathcal{E}\},
    \qquad
    d_i:=|\mathcal{E}_i|.
\]
The second layer sums the sampled wins of anchor $i$ and thresholds the result:
\begin{equation}
    \Gamma_i^{\step,k}(x)
    =
    H\left(
        \sum_{j\in\mathcal{E}_i}\Phi_{ij}^{\step}(x)
        -d_i+k-\frac{1}{2}
    \right).
    \label{eq:app-step-rvn-cell}
\end{equation}
Equivalently, if
\[
    L_i(x;\mathcal{E}_i)
    :=
    \sum_{j\in\mathcal{E}_i}
    \1\{\|x-x_j\|_2^2<\|x-x_i\|_2^2\}
\]
denotes the number of sampled rivals that are strictly closer to $x$ than $x_i$, then, away from distance ties and $k\ge 1$,
\begin{equation}
    \Gamma_i^{\step,k}(x)
    =
    \1\{L_i(x;\mathcal{E}_i)\le k-1\}.
    \label{eq:app-step-rvn-loss-count}
\end{equation}
Hence a step cell is active exactly when its anchor loses to at most $k-1$
sampled rivals. For $k=1$, this is the sampled hard Voronoi-cell condition.
For general $k$, it is a sampled top-$k$ condition.

The canonical $k$-NN readout attaches the target value $t_i$ to the cell for
anchor $i$:
\begin{equation}
    W^{(3),\step,k}_{:,i}
    =
    \frac{1}{k}t_i,
    \qquad
    b^{(3),\step,k}=0,
\end{equation}
which gives
\begin{equation}
    f^{\step,k}_{A,\mathcal{E}}(x)
    =
    \frac{1}{k}
    \sum_{i\in A}
    \Gamma_i^{\step,k}(x)t_i.
    \label{eq:app-step-rvn-readout}
\end{equation}
When the comparison graph is exactly $\mathcal E_A=\{(i,j):i,j\in A,\ i\neq j\}$, $1\le k\le M_C$, and distances to the selected anchors are pairwise distinct, exactly the anchors whose distance rank among $A$ is at most $k$ are active. Therefore,
\begin{equation}
    f^{\step,k}_{A,\mathcal{E}_{\mathrm{A}}}(x)
    =
    \frac{1}{k}
    \sum_{i\in\mathcal{N}_k(x;A)}
    t_i,
    \label{eq:app-step-rvn-selected-knn}
\end{equation}
where $\mathcal{N}_k(x;A)$ denotes the $k$ nearest anchors to $x$ among the
selected anchor set $A$. In particular, if $A=[N]$ and all ordered comparisons
are included, the step RVN exactly realizes the Euclidean $k$-nearest-neighbor
predictor on $\mathcal{D}_N$.

Allowing arbitrary readout parameters $\Theta=(W^{(3)},b^{(3)})$ gives the full
hypothesis class in \eqref{eq:rvn-hypothesis-class}. The canonical readout above
is the particular member that recovers the usual $k$-NN averaging rule.

\subsubsection{ReLU RVN}
\label{subsec:relu-rvn}

The ReLU RVN is a soft-cell relaxation of the hard Voronoi rule. Instead of
recording whether a comparison is won or lost, the first layer records the
amount by which the comparison for anchor $i$ is violated. For each ordered pair
$(i,j)\in\mathcal{E}$, define
\begin{equation}
    \Phi_{ij}^{\relu}(x)
    =
    \relu\!\left(
        \|x-x_i\|_2^2-\|x-x_j\|_2^2
    \right).
    \label{eq:app-relu-rvn-feature}
\end{equation}
Then $\Phi_{ij}^{\relu}(x)=0$ whenever $x_i$ is at least as close to $x$ as
$x_j$ is, and it is positive precisely when $x_j$ beats $x_i$. The value of the
feature is the nonnegative part of the signed violation margin.

For each anchor $i\in A$, the accumulated sampled violation is
\begin{equation}
    V_i(x)
    :=
    \sum_{j\in\mathcal{E}_i}
    \Phi_{ij}^{\relu}(x).
    \label{eq:app-relu-rvn-violation}
\end{equation}
Thus $V_i(x)=0$ when all sampled Voronoi inequalities for anchor $i$ are
satisfied, and $V_i(x)$ increases as the query moves outside the sampled cell of
anchor $i$. With gate width $\varepsilon>0$, the ReLU cell activation is
\begin{equation}
    \Gamma_i^{\relu,\varepsilon}(x)
    =
    \relu\left(
        1-\frac{V_i(x)}{\varepsilon}
    \right).
    \label{eq:app-relu-rvn-cell}
\end{equation}
This cell equals one when the sampled comparisons for anchor $i$ are all
satisfied, decays linearly for accumulated violations between $0$ and
$\varepsilon$, and vanishes once $V_i(x)\ge\varepsilon$.

The ReLU RVN readout is affine:
\begin{equation}
    f_{A,\mathcal{E},\Theta}^{\relu,\varepsilon}(x)
    =
    W^{(3)}\Gamma_{A,\mathcal{E}}^{\relu,\varepsilon}(x)+b^{(3)}.
    \label{eq:app-relu-rvn-readout}
\end{equation}
The canonical nearest-neighbor readout uses
\[
    W^{(3),\relu}_{:,i}=t_i,
    \qquad
    b^{(3),\relu}=0,
\]
so that
\begin{equation}
    f_{A,\mathcal{E}}^{\relu,\varepsilon}(x)
    =
    \sum_{i\in A}
    \Gamma_i^{\relu,\varepsilon}(x)t_i.
    \label{eq:app-relu-rvn-canonical-readout}
\end{equation}

When the comparison graph is complete over $A$ with $\mathcal E_A=\{(i,j):i,j\in A,\ i\neq j\}$, the only anchor with zero
accumulated violation is the nearest anchor, provided there are no distance
ties. Hence, as $\varepsilon\to0^+$, the ReLU RVN selects the same cell as the
hard $1$-nearest-neighbor rule:
\begin{equation}
    \lim_{\varepsilon\to0^+}
    \Gamma_i^{\relu,\varepsilon}(x)
    =
    \1\{i=\argmin_{\ell\in A}\|x-x_\ell\|_2^2\},
    \label{eq:app-relu-rvn-limit-cell}
\end{equation}
again away from ties. Consequently, with the canonical readout,
\begin{equation}
    \lim_{\varepsilon\to0^+}
    f_{A,\mathcal{E}_{\mathrm{A}}}^{\relu,\varepsilon}(x)
    =
    t_{\argmin_{i\in A}\|x-x_i\|_2^2}.
    \label{eq:app-relu-rvn-1nn-limit}
\end{equation}
If $A=[N]$, this is exactly the Euclidean $1$-nearest-neighbor predictor on the
full reference dataset.

Thus the ReLU RVN is a standard three-layer ReLU MLP realization of Voronoi
geometry: the first hidden layer computes violation margins for sampled
bisectors, the second hidden layer converts accumulated violations into soft
cell activations, and the readout layer combines the targets or logits attached
to the active cells. Unlike the step RVN, which directly realizes hard top-$k$
selection, the ReLU construction is most naturally interpreted as a soft
nearest-cell relaxation whose zero-width limit recovers $1$-NN.

\section{Proofs of the main statements}\label{app:proofs}

\subsection{Proof of Theorem~\ref{thm:step-rvn-knn-equivalence}}
\label{app:proof-step-rvn-knn-limit}

\noindent\textbf{Theorem~\ref{thm:step-rvn-knn-equivalence} (restated).}
Fix a reference dataset
$\mathcal{D}_N=\{(x_i,t_i)\}_{i=1}^N$ and $1\le k\le N$.
If $A=[N]$ and
\[
\mathcal{E}_{\mathrm{full}}
=
\{(i,j):i,j\in[N],\ i\neq j\},
\]
then the step RVN with canonical readout
$W^{(3)}_{:,i}=t_i/k$ and $b^{(3)}=0$ satisfies
\[
f_{\mathrm{RVN}}^{\step,k}(x)
=
f_{k\mathrm{NN},\mathcal{D}_N}(x)
\]
for every $x$ with no distance ties.

\begin{proof}
Fix $\Omega_X\subset\R^d$ denoting the set of query points with no distance ties
among the reference samples, and take
$A=[N]$ and $\mathcal{E}=\mathcal{E}_{\mathrm{full}}$.
If $N=1$, necessarily $k=1$ and there are no comparisons.
The step-cell activation is
$\Gamma_1^{\step,1}(x)=\sigma_H(1/2)=1$,
so the result is immediate. Henceforth assume $N\ge2$.

Let $r_i(x)$ denote the Euclidean distance rank of $x_i$ among
$x_1,\ldots,x_N$, with $r_i(x)=1$ for the nearest reference sample. Because
$x\in\Omega_X$, there are no distance ties. For the step Voronoi feature,
\[
    \Phi_{ij}^{\step}(x)
    =
    \1\{\|x-x_i\|_2^2\le \|x-x_j\|_2^2\}.
\]
Hence $x_i$ beats exactly the $N-r_i(x)$ reference samples that are farther
from $x$ than $x_i$ is:
\begin{equation}
    \sum_{j\in[N]:j\neq i}\Phi_{ij}^{\step}(x)
    =
    N-r_i(x).
    \label{eq:app-step-rvn-wins}
\end{equation}
In the full-comparison regime, $d_i=N-1$. Substituting
\eqref{eq:app-step-rvn-wins} into the step cell layer yields
\begin{align}
    \Gamma_i^{\step,k}(x)
    &=
    \sigma_H\left(
        \sum_{j\neq i}\Phi_{ij}^{\step}(x)
        -d_i+k-\frac12
    \right) \\
    &=
    \sigma_H\left(
        N-r_i(x)-(N-1)+k-\frac12
    \right) \\
    &=
    \sigma_H\left(
        k+\frac12-r_i(x)
    \right).
\end{align}
Since $r_i(x)$ is integer-valued,
\begin{equation}
    \Gamma_i^{\step,k}(x)=1
    \quad\Longleftrightarrow\quad
    r_i(x)\le k.
    \label{eq:app-step-rvn-active-cells}
\end{equation}
Therefore, the active step RVN cells are exactly the $k$ nearest neighbors of
$x$.

With readout
\[
    W^{(3),\step,k}_{:,i}=\frac{t_i}{k},
    \qquad
    b^{(3),\step,k}=0,
\]
the RVN output is
\begin{align}
    f_{\mathrm{RVN}}^{\step,k}(x)
    &=
    \frac{1}{k}\sum_{i=1}^N \Gamma_i^{\step,k}(x)t_i \\
    &=
    \frac{1}{k}\sum_{i\in\mathcal{N}_k(x;X)} t_i \\
    &=
    f_{k\mathrm{NN},\mathcal{D}_N}(x).
\end{align}
This proves the theorem.
\end{proof}

\subsection{Proof of Theorem~\ref{thm:relu-rvn-1nn-equivalence}}
\label{app:proof-relu-rvn-1nn-limit}

\noindent\textbf{Theorem~\ref{thm:relu-rvn-1nn-equivalence} (restated).}
Fix a reference dataset $\mathcal{D}_N=\{(x_i,t_i)\}_{i=1}^N$ with $N\ge1$.
Suppose $A=[N]$ and
\[
    \mathcal{E}
    =
    \mathcal{E}_{\mathrm{full}}
    =
    \{(i,j):i,j\in[N],\ i\neq j\},
\]
and use the canonical readout
$W^{(3)}_{:,i}=t_i$, $b^{(3)}=0$.

Fix $x\in\Omega_X$. For $N\ge2$, let
\[
    \Delta(x)
    =
    \|x-x_{i_2(x)}\|_2^2-\|x-x_{i_1(x)}\|_2^2
\]
be the squared-distance gap between the nearest and second-nearest
reference samples. Then
\[
    f_{\mathrm{RVN}}^{\relu,\varepsilon}(x)
    =
    f_{1\mathrm{NN},\mathcal{D}_N}(x)
    \qquad
    \text{for every }0<\varepsilon\le\Delta(x).
\]
If $N=1$, this equality holds for every $\varepsilon>0$.
Consequently, for every $x\in\Omega_X$,
\[
    \lim_{\varepsilon\to0^+}
    f_{\mathrm{RVN}}^{\relu,\varepsilon}(x)
    =
    f_{1\mathrm{NN},\mathcal{D}_N}(x).
\]

\begin{proof}
Fix $x\in\Omega_X$, and take
$A=[N]$ and $\mathcal{E}=\mathcal{E}_{\mathrm{full}}$.

If $N=1$, there are no comparisons, so $V_1(x)=0$ and $\Gamma_1^{\relu,\varepsilon}(x)=1$ for every $\varepsilon>0$. The canonical readout therefore gives $f_{\mathrm{RVN}}^{\relu,\varepsilon}(x)=t_1=f_{1\mathrm{NN},\mathcal{D}_1}(x)$. Henceforth assume $N\ge2$.

Let
\[
    i^\star(x)
    =
    \argmin_{i\in[N]}\|x-x_i\|_2^2
\]
be the unique nearest neighbor of $x$. The ReLU Voronoi feature records the
amount by which the comparison for anchor $i$ is violated:
\[
\begin{aligned}
    \Phi_{ij}^{\relu}(x)&=\relu(-h_{ij}(x)), \\
    h_{ij}(x)&=\|x-x_j\|_2^2-\|x-x_i\|_2^2.
\end{aligned}
\]
For the nearest anchor $i^\star(x)$,
\[
    \|x-x_{i^\star}\|_2^2
    <
    \|x-x_j\|_2^2
    \qquad
    \text{for all } j\neq i^\star.
\]
Therefore $h_{i^\star j}(x)>0$ and
\[
    \Phi_{i^\star j}^{\relu}(x)=0
    \qquad
    \text{for all } j\neq i^\star.
\]
Its accumulated violation is therefore
\begin{equation}
    V_{i^\star}(x)
    =
    \sum_{j\neq i^\star}\Phi_{i^\star j}^{\relu}(x)
    =
    0.
    \label{eq:app-relu-nearest-zero-violation}
\end{equation}

Now take any $i\neq i^\star(x)$. Since the comparison graph is complete, the
comparison $(i,i^\star)$ is present. Moreover,
\[
    h_{i i^\star}(x)
    =
    \|x-x_{i^\star}\|_2^2-\|x-x_i\|_2^2
    <
    0.
\]
Hence
\[
    \Phi_{i i^\star}^{\relu}(x)
    =
    \|x-x_i\|_2^2-\|x-x_{i^\star}\|_2^2
    >
    0.
\]
Thus every non-nearest anchor has strictly positive accumulated violation:
\begin{equation}
    V_i(x)
    =
    \sum_{j\neq i}\Phi_{ij}^{\relu}(x)
    \ge
    \Phi_{i i^\star}^{\relu}(x)
    >
    0,
    \qquad i\neq i^\star(x).
    \label{eq:app-relu-nonnearest-positive-violation}
\end{equation}

Since $i_2(x)$ is the second-nearest reference sample, for every
$i\neq i^\star(x)$ we have
\[
V_i(x)\ge \Phi_{i i^\star}^{\relu}(x)
=\|x-x_i\|_2^2-\|x-x_{i^\star}\|_2^2
\ge \Delta(x).
\]
The second-nearest anchor loses only to the nearest anchor, so
$V_{i_2(x)}(x)=\Delta(x)$. Thus, for every
$0<\varepsilon\le\Delta(x)$, the ReLU gate gives
$\Gamma_{i^\star}^{\relu,\varepsilon}(x)=1$ and
$\Gamma_i^{\relu,\varepsilon}(x)=0$ for every $i\neq i^\star(x)$.
The canonical readout therefore satisfies
\[
f_{\mathrm{RVN}}^{\relu,\varepsilon}(x)
=t_{i^\star(x)}
=f_{1\mathrm{NN},\mathcal{D}_N}(x).
\]
Since $\Delta(x)>0$, the limit as $\varepsilon\to0^+$ follows.

\end{proof}

\section{Budgeted Approximation: Proofs}
\label{app:budgeted-approximation-proofs}

\subsection{Sampled comparisons control top-\texorpdfstring{$k$}{k} activation}

\noindent\textbf{Proposition~\ref{prop:rvn-hypergeom} (restated).}
For a main anchor $i\in A$, sample
$\mathcal{E}_i\subseteq[N]\setminus\{i\}$ uniformly without replacement with
$|\mathcal{E}_i|=H$. If $r_i^X(x)$ is the full-reference rank defined in
Equation~\eqref{eq:rvn-reference-rank}, then
\[
L_i(x;\mathcal{E}_i)
\sim
\operatorname{Hypergeom}(N-1,r_i^X(x)-1,H),
\]
and $\Gamma_i^{\step,k}(x)=1$ exactly when
$L_i(x;\mathcal{E}_i)\le k-1$.

\begin{proof}
Fix an input $x\in\R^d$ and a main anchor $i$. Write
$r:=r_i^X(x)$. By definition of full-reference rank, exactly $r-1$ of the
$N-1$ possible rivals are strictly closer to $x$ than $x_i$, while the remaining
$N-r$ rivals are at least as far from $x$ as $x_i$.

The set $\mathcal{E}_i$ is drawn uniformly from all
$\binom{N-1}{H}$ subsets of $H$ rivals. The loss count
$L_i(x;\mathcal{E}_i)$ records how many of the sampled rivals come from the
$r-1$ closer reference points. To obtain exactly $\ell$ losses, the sample must
contain $\ell$ of those $r-1$ closer points and $H-\ell$ of the $N-r$ non-closer
points. The number of such rival sets is therefore
\[
\binom{r-1}{\ell}\binom{N-r}{H-\ell}.
\]
Dividing by the total number of size-$H$ rival sets gives
\begin{equation}
\Prob\!\left(L_i(x;\mathcal{E}_i)=\ell\right)
=
\frac{\binom{r_i^X(x)-1}{\ell}
      \binom{N-r_i^X(x)}{H-\ell}}
     {\binom{N-1}{H}},
\label{eq:app-rvn-hypergeom-pmf}
\end{equation}
with infeasible binomial coefficients interpreted as zero. This is exactly the
probability mass function of
$\operatorname{Hypergeom}(N-1,r_i^X(x)-1,H)$.

By Equation~\eqref{eq:rvn-sampled-loss-count} and the step-cell rule,
$\Gamma_i^{\step,k}(x)=1$ iff $L_i(x;\mathcal{E}_i)\le k-1$. Hence
\begin{equation}
\Prob\!\left(\Gamma_i^{\step,k}(x)=1\right)
=
\sum_{\ell=0}^{\min\{k-1,H\}}
\frac{\binom{r_i^X(x)-1}{\ell}
      \binom{N-r_i^X(x)}{H-\ell}}
     {\binom{N-1}{H}}.
\label{eq:app-rvn-hypergeom-activation}
\end{equation}
If $r_i^X(x)\le k$, at most $k-1$ closer reference points exist, so the cell is
active for every possible rival sample. At the complete rival budget
$H=N-1$, all possible rivals are observed and
$L_i(x;\mathcal{E}_i)=r_i^X(x)-1$ deterministically; therefore the selected
main anchor is active exactly when its global reference rank is at most $k$.
\end{proof}

\subsection{Comparison budget controls effective neighborhood size}

\noindent\textbf{Corollary~\ref{cor:rvn-effective-neighborhood} (restated).}
Fix $k\ge 1$, $1\le M_C\le N$, and $0\le H\le N-1$. Suppose $A$ is sampled uniformly from $[N]$ with $|A|=M_C$, and for each
$i\in A$, $H$ rivals are sampled uniformly from $[N]\setminus\{i\}$. Assume that the reference samples have pairwise distinct distances to the query $x$. Then
\[
\E[S_H^k(x)]
=M_C\frac{\min\{k,H+1\}}{H+1}.
\]
For $H\ge k-1$,
$k_{\mathrm{eff}}=\E[S_H^k(x)]=M_Ck/(H+1)=M_C^2k/(M_F+M_C)$.

\begin{proof}
After sampling the main-anchor set $A$, let $I$ be an index chosen uniformly
from the $M_C$ elements of $A$. Since $A$ itself is a uniformly sampled
size-$M_C$ subset of $[N]$, $I$ is marginally uniform on the full reference set:
\[
\Prob(I=i)
=\Prob(i\in A)\Prob(I=i\mid i\in A)
=\frac{M_C}{N}\frac{1}{M_C}
=\frac{1}{N}.
\]
Conditional on $I=i$, its rival set $\mathcal{E}_I$ is a uniformly sampled
size-$H$ subset of $[N]\setminus\{i\}$.

Let $B:=\{I\}\cup\mathcal{E}_I$, so $|B|=H+1$. For any fixed subset
$B\subset[N]$ of size $H+1$ and any $i\in B$,
\[
\Prob(B, I=i)
=\frac{1}{N}\frac{1}{\binom{N-1}{H}}
=\frac{1}{\binom{N}{H+1}}\frac{1}{H+1},
\]
where the second equality follows from
$\binom{N}{H+1}=\frac{N}{H+1}\binom{N-1}{H}$. Thus $B$ is uniformly distributed
over all $(H+1)$-subsets of the reference set and, conditional on $B$, the
distinguished main anchor $I$ is uniform over the $H+1$ elements of $B$.

Order the elements of $B$ by increasing distance to $x$. Away from ties, one
element occupies each relative rank in $\{1,\ldots,H+1\}$. Because $I$ is
uniform on $B$, its relative rank $R_B(I)$ is uniform:
\[
\Prob(R_B(I)=r\mid B)=\frac{1}{H+1},
\qquad r=1,\ldots,H+1.
\]
The sampled loss count is exactly $R_B(I)-1$, since every element of $B$ ranked
before $I$ is one sampled rival closer to $x$. Therefore
\[
\begin{aligned}
\Gamma_I^{\step,k}(x)=1
&\iff R_B(I)-1 \le k-1 \\
&\iff R_B(I)\le k.
\end{aligned}
\]
There are $\min\{k,H+1\}$ admissible ranks among the $H+1$ equally likely
possibilities, and hence
\begin{equation}
\begin{aligned}
\Prob\!\left(\Gamma_I^{\step,k}(x)=1\right)
&= \sum_{r=1}^{\min\{k,H+1\}} \frac{1}{H+1} \\
&= \frac{\min\{k,H+1\}}{H+1}.
\end{aligned}
\label{eq:app-rvn-effective-probability}
\end{equation}

Finally, because $I$ is uniform among the selected main anchors,
\[
\begin{aligned}
\Prob\!\left(\Gamma_I^{\step,k}(x)=1\right)
&= \E\!\left[
    \frac{1}{M_C}
    \sum_{i\in A}\Gamma_i^{\step,k}(x)
   \right] \\
&= \frac{1}{M_C}\E[S_H^k(x)].
\end{aligned}
\]
Multiplying by $M_C$ proves
\[
\E[S_H^k(x)]
=M_C\frac{\min\{k,H+1\}}{H+1}.
\]
For $H\ge k-1$, the minimum equals $k$. Substituting $M_F=M_CH$ yields
\[
k_{\mathrm{eff}}
=\E[S_H^k(x)]
=\frac{M_Ck}{H+1}
=\frac{M_C^2k}{M_F+M_C}.
\]
At $H=N-1$, this becomes $k_{\mathrm{eff}}=M_Ck/N$; when also $M_C=N$, it
reduces to the exact $k$ active cells of full $k$-NN.
\end{proof}

\subsection{Complete RVFs encode Kendall rank geometry}

\noindent\textbf{Proposition~\ref{prop:rvf-kendall-kernel} (restated).}
Assume $N\ge 2$ and $x,x'\in\Omega_X$. Let $P=\binom{N}{2}$ and
$K_X(x,x')=P^{-1}\sum_{i<j}\psi_{ij}(x)\psi_{ij}(x')$. Then
\[
K_X(x,x')
=1-\frac{2D_K(\pi_X(x),\pi_X(x'))}{P},
\]
so $K_X$ equals Kendall rank correlation and is positive semidefinite.

\begin{proof}
For an unordered reference pair $\{i,j\}$, call the pair \emph{concordant} if
$x$ and $x'$ rank $i$ and $j$ in the same order, and \emph{discordant} if the
order is reversed. Equivalently,
$\psi_{ij}(x)\psi_{ij}(x')=+1$ for a concordant pair and $-1$ for a discordant
pair. Let $C$ and $D$ denote the numbers of concordant and discordant pairs.
Then $C+D=P$, and the Kendall disagreement count is
$D_K(\pi_X(x),\pi_X(x'))=D$. Therefore
\[
\begin{aligned}
K_X(x,x')
&= \frac{C-D}{P}
 = \frac{P-2D}{P} \\
&= 1-\frac{2D_K(\pi_X(x),\pi_X(x'))}{P}.
\end{aligned}
\]
This is Kendall rank correlation. Moreover,
$K_X(x,x')=z_X(x)^\top z_X(x')$ for the explicit feature map
$z_X(x)=P^{-1/2}(\psi_{ij}(x))_{i<j}$, proving positive semidefiniteness.
\end{proof}

\subsection{Ranking information retained by ReLU RVFs}
\label{app:relu-ranking}

For inputs without distance ties,
$\Phi_{ij}^{\relu}(x)=\max\{0,-h_{ij}(x)\}$ is positive exactly when
reference sample $j$ is closer than $i$. Therefore,
\[
\psi_{ij}(x)
=1-2\cdot \mathbf{1}\{\Phi_{ij}^{\relu}(x)>0\}.
\]
Thus, thresholding the complete ReLU-RVF map recovers the signed
comparisons and hence the Kendall kernel in
Proposition~\ref{prop:rvf-kendall-kernel}.
The concentration bound also applies to these thresholded features
under independent uniform pair sampling, but does not extend directly
to raw ReLU inner products which also contain information regarding magnitude.
\citet{10.1007/978-3-642-03784-9_34} established a Hoeffding bound for
estimating Kendall's rank correlation between two rankings by uniformly
sampling comparison pairs. A union bound gives the following guarantee
over a fixed set of queries.

\subsection{Random RVFs preserve rank geometry with logarithmic feature growth}

\noindent\textbf{Corollary~\ref{cor:rvf-kendall-concentration} (restated).}
Assume $N\ge2$, and let $\mathcal Q=\{z_1,\ldots,z_m\}\subset\Omega_X$ be a fixed set of $m\ge1$ queries. If $M_F$ canonical
reference pairs are sampled independently and uniformly, then
$\widehat K_{M_F}$ is unbiased for $K_X$. All pairwise kernel values on
$\mathcal{Q}$ are within $\epsilon$ of their complete values with probability
at least $1-\delta$ whenever
\[
M_F\ge \frac{2}{\epsilon^2}\log\frac{2m^2}{\delta}.
\]

\begin{proof}
For a uniformly sampled canonical reference pair $e$, define
$Y_e:=\psi_e(x)\psi_e(x')\in\{-1,+1\}$. Uniformity over the
$P=\binom{N}{2}$ reference pairs gives
\[
\E[Y_e]
=\frac{1}{P}\sum_{i<j}\psi_{ij}(x)\psi_{ij}(x')
=K_X(x,x').
\]
Hence $\widehat K_{M_F}=M_F^{-1}\sum_{q=1}^{M_F}Y_{e_q}$ is unbiased. Since the
sampled pairs are independent and each $Y_{e_q}\in[-1,1]$, Hoeffding's
inequality yields
\begin{equation}
\Prob\!\left(
|\widehat K_{M_F}(x,x')-K_X(x,x')|\ge\epsilon
\right)
\le 2e^{-M_F\epsilon^2/2}.
\label{eq:app-rvf-kendall-concentration}
\end{equation}
Apply this bound to the at most $m^2$ ordered query pairs
$(z_a,z_b)\in\mathcal{Q}\times\mathcal{Q}$ and use the union bound:
\[
\begin{aligned}
\Prob\!\Bigl[
\max_{a,b\in[m]}
\bigl|\widehat K_{M_F}(z_a,z_b)-K_X(z_a,z_b)\bigr|
&\ge \epsilon
\Bigr] \\
&\le 2m^2 e^{-M_F\epsilon^2/2}.
\end{aligned}
\]
Requiring the right-hand side to be at most $\delta$ gives
\[
M_F\ge \frac{2}{\epsilon^2}\log\frac{2m^2}{\delta},
\]
which proves the simultaneous approximation guarantee. Equivalently,
$M_F=O(\epsilon^{-2}[\log m+\log(1/\delta)])$ suffices.
\end{proof}

\section{Selector Algorithms}
\label{appendix:selectors}

The selectors in Section~\ref{subsec:rvf-selectors} are defined in Algorithms~\ref{alg:random-selector}, \ref{alg:most-diverse-selector}, \ref{alg:sparse-dimension-selector}, and \ref{alg:variance-selector}.

For the Cosine Diversity selector (assuming $d\ge 1$, $H\ge 2$, and distinct reference locations), $\tau$ is computed for each dataset based on the ambient dimension $d$ and group size $H$. Using a Gaussian approximation for pairwise cosine similarities, we set a Bonferroni-corrected baseline for the $m=H(H-1)/2$ comparisons at nominal level $\alpha=0.1$:
$
\tau_{\text{stat}} = \frac{1}{\sqrt{d}}\Phi^{-1}\left(1 - \frac{\alpha}{2m}\right).
$
To account for dimensional constraints when $H>d$, we also use the Welch coherence lower bound
$
    \mu_{\min} = \sqrt{\frac{\max(0, H-d)}{d(H-1)}}.
$
Finally, we take $\tau = \max\left(\tau_{\text{stat}}, \, \mu_{\min} + \epsilon\right)$ with safety margin $\epsilon = 0.05$, and clip $\tau$ between $0.1$ and $0.9$ to prevent overly restrictive or permissive selection. 
This threshold is a heuristic: the Welch bound provides a necessary lower bound on coherence rather than a guarantee that a feasible subset exists among the candidate directions. If the greedy scan selects fewer than $H$ rivals, the selector samples the remaining slots uniformly without replacement from the unused rivals. This gives exactly $H$ rivals whenever $H\le N-1$.

\begin{algorithm}[htbp]
\caption{Random selector}
\label{alg:random-selector}
\begin{algorithmic}[1]
\Require Dataset $\{x_i\}_{i=1}^N$, Number of features $M_F$, Number of cells $M_C$
\State Set Number of features per cell $H \gets M_F/M_C$
\State Sample main anchors $A \subseteq [N]$ uniformly with $|A|=M_C$
\State Initialize selected pair set $\mathcal{E}\gets \emptyset$
\For{$i \in A$}
    \State Sample $S_i \subseteq [N]\setminus\{i\}$ uniformly with $|S_i|=H$
    \State Add $\{(i,j): j\in S_i\}$ to $\mathcal{E}$
\EndFor
\State \Return $\mathcal{E}$
\end{algorithmic}
\end{algorithm}

\begin{algorithm}[htbp]
\caption{Cosine Diversity selector}
\label{alg:most-diverse-selector}
\begin{algorithmic}[1]
\Require Distinct dataset $\{x_i\}_{i=1}^N$, Number of features $M_F$, Number of cells $M_C$, threshold $\tau$
\State Set Number of features per cell $H \gets M_F/M_C$
\State Sample main anchors $A \subseteq [N]$ uniformly with $|A|=M_C$
\State Initialize selected pair set $\mathcal{E}\gets \emptyset$
\For{$i \in A$}
    \State Initialize selected neighbor set $S_i \gets \emptyset$
    \State Randomly order candidates $C_i \gets [N]\setminus\{i\}$
    \For{$j \in C_i$}
        \State Compute $u_{ij} \gets (x_j-x_i)/\|x_j-x_i\|_2$
        \If{$|S_i| < H$ and ($S_i=\emptyset$ or $\max_{\ell\in S_i}|\langle u_{ij},u_{i\ell}\rangle| < \tau$)}
            \State Add $j$ to $S_i$
        \EndIf
        \If{$|S_i|=H$}
            \State \textbf{break}
        \EndIf
    \EndFor
    \If{$|S_i|<H$}
        \State Fill $S_i$ with uniformly sampled remaining candidates until $|S_i|=H$
    \EndIf
    \State Add $\{(i,j): j\in S_i\}$ to $\mathcal{E}$
\EndFor
\State \Return $\mathcal{E}$
\end{algorithmic}
\end{algorithm}

\begin{algorithm}[htbp]
\caption{Sparse Dimension selector}
\label{alg:sparse-dimension-selector}
\begin{algorithmic}[1]
\Require Dataset $\{x_i\}_{i=1}^N$, Number of features $M_F$, Number of cells $M_C$
\State Set Number of features per cell $H \gets M_F/M_C$
\State Sample main anchors $A \subseteq [N]$ uniformly with $|A|=M_C$
\State Initialize selected pair set $\mathcal{E}\gets \emptyset$
\For{$i \in A$}
    \For{$j \in [N]\setminus\{i\}$}
        \State Compute $d_{ij}\gets |x_i-x_j|$
        \State Compute sparsity score $q_{ij}\gets \operatorname{sp}(d_{ij})$ via \eqref{eq:hoyer-sparsity}
    \EndFor
    \State Select $S_i$ as the $H$ candidates with largest $q_{ij}$
    \State Add $\{(i,j): j\in S_i\}$ to $\mathcal{E}$
\EndFor
\State \Return $\mathcal{E}$
\end{algorithmic}
\end{algorithm}

\begin{algorithm}[htbp]
\caption{High Variance selector}
\label{alg:variance-selector}
\begin{algorithmic}[1]
\Require Dataset $\{x_i\}_{i=1}^N$, Number of features $M_F$, Number of cells $M_C$, nonempty evaluation subset $\mathcal{Z}$
\State Set Number of features per cell $H \gets M_F/M_C$
\State Sample main anchors $A \subseteq [N]$ uniformly with $|A|=M_C$
\State Initialize selected pair set $\mathcal{E}\gets \emptyset$
\For{$i \in A$}
    \For{$j \in [N]\setminus\{i\}$}
        \For{$z \in \mathcal{Z}$}
            \State Compute
            \[
                \chi_{ij}(z)
                \gets
                \1\{\|z-x_i\|_2^2 \le \|z-x_j\|_2^2\}
            \]
        \EndFor
        \State Estimate $\widehat p_{ij}\gets |\mathcal{Z}|^{-1}\sum_{z\in\mathcal{Z}}\chi_{ij}(z)$
        \State Score candidate $j$ by $q_{ij}\gets \widehat p_{ij}(1-\widehat p_{ij})$
    \EndFor
    \State Select $S_i$ as the $H$ candidates with largest $q_{ij}$
    \State Add $\{(i,j): j\in S_i\}$ to $\mathcal{E}$
\EndFor
\State \Return $\mathcal{E}$
\end{algorithmic}
\end{algorithm}

\section{Additional experimental details}
\label{appendix:rvf-supervised-details}

\subsection{Individual Dataset Results}
\label{appendix:individual_datasets}
\begin{figure}[t]
    \centering
    \includegraphics[width=\linewidth]{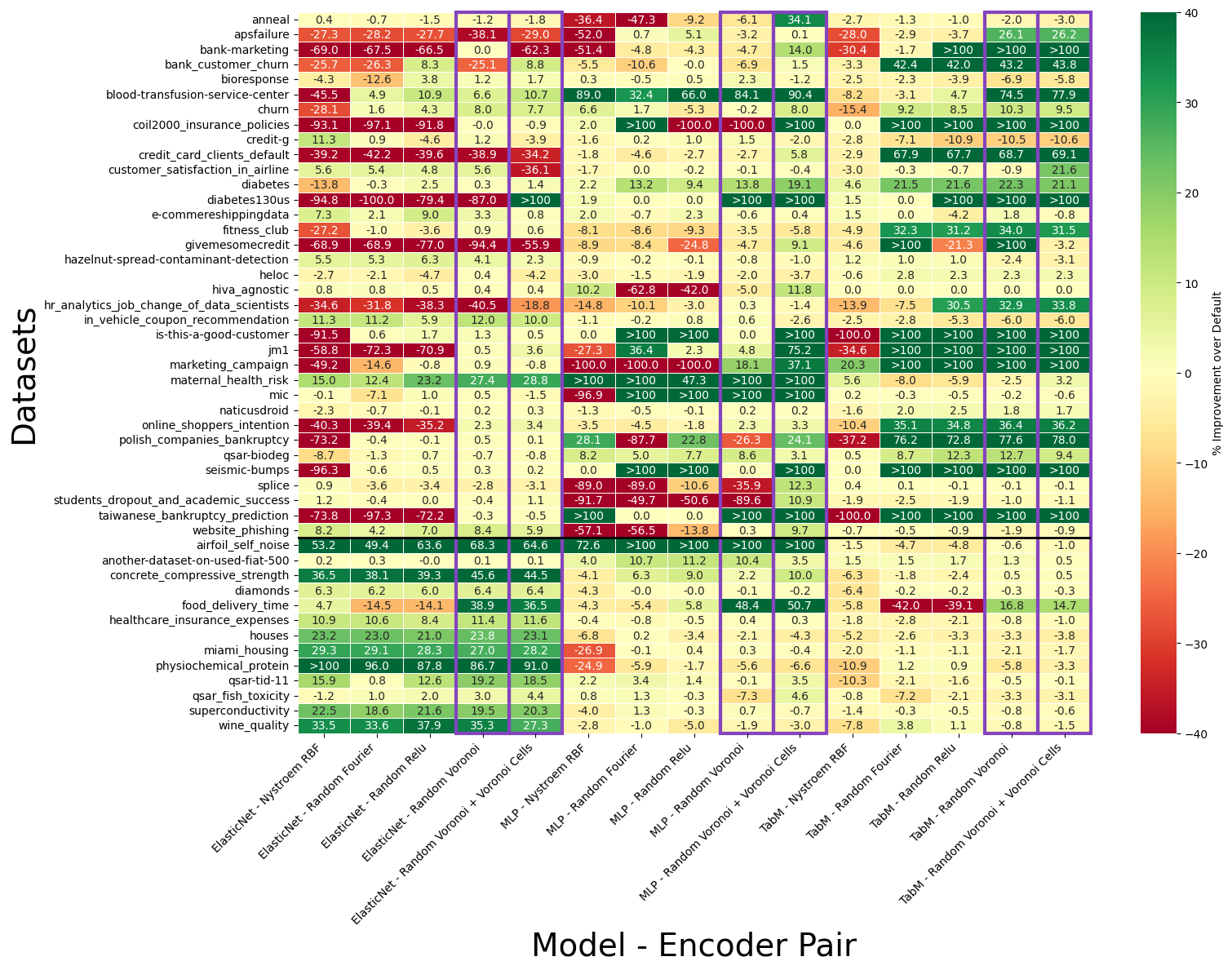}
    \caption{\textbf{Individual Dataset Results. } Percentage improvement for each dataset over all models and encoders compared to raw (non-encoded) baseline. Classification measured using F1 score (above black line) and regression measured using $R^2$ score (below black line). RVF Based encoders perform better than other encoders for most datasets within the same model group. These results are especially pronounced for Modern Tabular Models like TabM. All classification datasets are above the black horizontal line, while regression tasks are below. }
    \label{fig:individual_datasets}
\end{figure}

\subsection{Classifier and encoder ablation}
\label{appendix:classifier-encoder-ablation}

To separate RVN's role as a nearest-neighbor classifier from RVF's role as a reusable encoder, we compare standard $k$-NN, the aggregated Voronoi Cell representation, the pairwise Voronoi Feature representation, and their combination with the same ElasticNet downstream model where applicable.

\begin{table}[htbp]
\centering
\caption{\textbf{Classifier-versus-encoder ablation.} Elo scores separate the nearest-neighbor/cell representation from the pairwise RVF representation.}
\label{tab:classifier-encoder-ablation}
\small
\begin{tabular}{lr}
\toprule
Configuration & Elo \\
\midrule
Voronoi Features + Cells + ElasticNet & 1142 \\
Voronoi Features + ElasticNet & 1140 \\
Raw + ElasticNet & 971 \\
Standard $k$-NN & 881 \\
Voronoi Cells + ElasticNet & 866 \\
\bottomrule
\end{tabular}
\end{table}

Table~\ref{tab:classifier-encoder-ablation} shows that the pairwise Voronoi Features account for most of the downstream gain: RVF+ElasticNet and RVF+VC+ElasticNet reach about $1140$ Elo, whereas standard $k$-NN and Voronoi Cells+ElasticNet remain below $900$. The small difference between RVF and RVF+VC suggests that the individual pairwise margins, rather than the aggregated cell representation alone, drive most of the improvement.

\subsection{Benchmark protocol}
\label{appendix:benchmark-protocol}

We evaluate supervised RVF encoders on TabArena-Lite~\citep{erickson2025tabarena}. The datasets used here range from roughly $500$ to $81{,}000$ samples and span classification and regression tasks, feature dimensions, and tabular domains. This is the practical fixed-budget evaluation of RVF and is distinct from the $N<1000$ full-RVN diagnostic in Section~\ref{subsec:rvf-1nn-comparison}. The goal is to test whether an encoder improves downstream learning across heterogeneous tabular problems, rather than on a single curated benchmark. Following the established benchmark protocol that is shown to be a good indicator of performance \citep{erickson2025tabarena}, we test on fold $0$. All RVF anchors, comparison pairs, scalers, and categorical encoders are fit using only the training split, avoiding test-set leakage. We encode all categorical features with one-hot encoding. 

For each dataset, we compare the following input encodings:
\begin{itemize}[leftmargin=1.6em]
    \item \textbf{None:} the raw input representation;
    \item \textbf{RFF:} Random Fourier Features, which use random sinusoidal projections to approximate shift-invariant kernels~\citep{rahimi2007random};
    \item \textbf{RRF:} Random ReLU Features, which use data-agnostic random affine projections followed by ReLU activations~\citep{hsu2021random};
    \item \textbf{RBF}: Nyström RBF features, which approximate data-dependent radial basis functions by random sampling~\citep{williamsnystromrbf}.  
    \item \textbf{RVF:} ReLU Random Voronoi Features $\Phi^{\relu}(x)$;
    \item \textbf{RVF+VC:} the concatenated representation $[\Phi^{\relu}(x);\Gamma^{\relu, \varepsilon}(x)]$.
\end{itemize}

We apply these encodings to the following models as downstream predictors: 
\begin{itemize}[leftmargin=1.6em]
    \item \textbf{ElasticNet:} a generalized linear model with combined $L_1$ and $L_2$ regularization, used to test whether the encoding exposes useful linear structure;
    \item \textbf{TorchMLP:} a standard multilayer perceptron, used to test interaction with generic neural predictors;
    \item \textbf{TabM-Mini:} a modern tabular neural model, used to test whether RVF improves specialized tabular architectures.
\end{itemize}

For additional neighborhood-model context, we report TabR~\citep{gorishniy2024tabr} and ModernNCA~\citep{ye2025modernnca} as aggregate Elo reference lines in Figure~\ref{fig:elo}. These models learn task-specific retrieval or neighborhood representations.

\subsection{Metrics and ranking}
\label{appendix:metrics-ranking}

To aggregate results over heterogeneous tasks, we adapt TabArena's Elo-based ranking methodology~\citep{elo1966uscf,erickson2025tabarena}, and use F1 for classification and $R^2$ for regression. Elo ratings are computed from pairwise comparisons induced by dataset-level task metrics, allowing classification and regression results to be summarized in a common ranking.
All Elo ratings are initialized at $1000$. 
We estimate uncertainty using $200$ bootstrap replicates, each sampling the original number of datasets with replacement and refitting Elo ratings.
We report the $2.5$th and $97.5$th percentiles of the resulting ratings as the $95\%$ bootstrap interval.

For Figure~\ref{fig:elo} (right), rankings are computed for each dataset--encoder pair for the same number of features. Concatenated RVF+VC features are adjusted accordingly to maintain the same size as the competitors to maintain fairness.  
As stated, RVF encodings do not increase their performance as TabR and ModernNCA already specialize for geometric nearest-neighbor bias. With RVF embeddings, TabR and ModernNCA get elo of 980 and 1020 respectively.

\subsection{Hyperparameter optimization}
\label{appendix:hpo_spaces}

\begin{table}[htbp]
\centering
\caption{Hyperparameter search space for ElasticNet.}
\label{table:hpo_elasticnet}
\begin{tabular}{ll}
\toprule
Hyperparameter & \multicolumn{1}{l}{Space} \\
\midrule
\texttt{alpha} & \multicolumn{1}{l}{LogUniform([1e-05, 1.0])} \\
\texttt{l1\_ratio} & \multicolumn{1}{l}{Uniform([0.0, 1.0])} \\
\texttt{scaler} & \multicolumn{1}{l}{Choice(["standard", "quantile"])} \\
\bottomrule
\end{tabular}
\end{table}

\begin{table}[htbp]
\centering
\caption{Hyperparameter search space for MLP.}
\label{table:hpo_mlp}
\begin{tabular}{ll}
\toprule
Hyperparameter & \multicolumn{1}{l}{Space} \\
\midrule
\texttt{hidden\_dim} & \multicolumn{1}{l}{UniformInt([8, 256])} \\
\texttt{activation} & \multicolumn{1}{l}{Choice(["relu", "elu"])} \\
\texttt{n\_layers} & \multicolumn{1}{l}{UniformInt([1, 5])} \\
\texttt{learning\_rate} & \multicolumn{1}{l}{LogUniform([0.0001, 0.03])} \\
\texttt{weight\_decay} & \multicolumn{1}{l}{LogUniform([1e-12, 0.1])} \\
\texttt{dropout\_rate} & \multicolumn{1}{l}{Uniform([0.0, 0.4])} \\
\texttt{use\_batch\_norm} & \multicolumn{1}{l}{Choice([True, False])} \\
\texttt{scaler} & \multicolumn{1}{l}{Choice(["standard", "quantile"])} \\
\bottomrule
\end{tabular}
\end{table}

\begin{table}[!t]
\centering
\caption{Hyperparameter search space for TabM.}
\label{table:hpo_tabm}
\begin{tabular}{lll}
\toprule
Hyperparameter & \multicolumn{2}{l}{Space} \\
\midrule
\texttt{arch\_type}
    & \multicolumn{2}{l}{Choice(["tabm-mini"])} \\
\texttt{tabm\_k}
    & \multicolumn{2}{l}{Choice([32])} \\
\shortstack[l]{\texttt{share\_training\_}\\\texttt{batches}}
    & \multicolumn{2}{l}{Choice([False])} \\
\texttt{batch\_size}
    & \multicolumn{2}{l}{Choice(["auto"])} \\
\texttt{patience}
    & \multicolumn{2}{l}{Choice([16])} \\
\texttt{lr}
    & \multicolumn{2}{l}{LogUniform([1e-4, 3e-3])} \\
\shortstack[l]{\texttt{weight\_decay\_}\\\texttt{choice}}
    & \multicolumn{2}{l}{Choice([0.0, 1.0])} \\
\shortstack[l]{\texttt{gradient\_clipping\_}\\\texttt{norm}}
    & \multicolumn{2}{l}{Choice([1.0])} \\
\texttt{amp}
    & \multicolumn{2}{l}{Choice([False])} \\
\texttt{compile\_model}
    & \multicolumn{2}{l}{Choice([False])} \\
\texttt{n\_blocks}
    & \multicolumn{2}{l}{Choice([2, 3, 4, 5])} \\
\texttt{d\_block}
    & \multicolumn{2}{l}{
        \shortstack[l]{Choice(128:16:1024)\\
        \textit{(step size = 16)}}
      } \\
\texttt{dropout\_choice}
    & \multicolumn{2}{l}{Choice([0.0, 1.0])} \\
\texttt{d\_embedding}
    & \multicolumn{2}{l}{
        \shortstack[l]{Choice([8, 12, 16, 20,\\
        24, 28, 32])}
      } \\
\texttt{num\_emb\_n\_bins}
    & \multicolumn{2}{l}{UniformInt([2, 128])} \\
\texttt{eval\_batch\_size}
    & \multicolumn{2}{l}{Choice([128])} \\
\texttt{bool\_to\_cat}
    & \multicolumn{2}{l}{Choice([True])} \\
\texttt{average\_logits}
    & \multicolumn{2}{l}{Choice([False])} \\
\bottomrule
\end{tabular}
\end{table}

For every dataset, encoder, and downstream model combination, we perform automated hyperparameter optimization with \texttt{Optuna}~\citep{akiba2019optuna} using the \texttt{TPESampler}. Each experiment is run for up to $50$ trials, optimizing over 1 fold validation set separate from test and train sets. We adopted the search spaces from \citet{erickson2025tabarena} for \texttt{torchMLP} and \texttt{TabM}. For the linear model with ElasticNet regularization, we used a custom wide search. Details of these are provided in Tables \ref{table:hpo_elasticnet}-\ref{table:hpo_tabm}. 

For the encoders, we perform HPO over number of features ranging from $5$ to $25000$. For RVF based encoders, $M_C$ ranges from $50$ to $500$, while also optimizing over the selectors, and choosing $\epsilon \in [10^{-5}, 10^5]$. For RFF and Nystroem RBF, the gamma parameter is optimized over the range of $0.0001$ to $100.0$. For all encoders, we perform standard scaling before and after encoding.

\subsection{Computational resources}
\label{appendix:compute-resources}

All experiments were run on cluster machines. The TabM and MLP benchmark experiments were run on NVIDIA A100 GPUs with 40GB of VRAM. All remaining experiments, including ElasticNet benchmarks, interpretability experiments, and $1$-NN/RVF equivalence experiments, were run on CPU. For these CPU experiments, each dataset--model--encoding--group configuration required less than 10GB of RAM.

\end{document}